\documentclass{article}

\usepackage{microtype}
\usepackage{graphicx}
\usepackage{subfigure}
\usepackage{booktabs} 

\usepackage{hyperref}

\usepackage[accepted]{icml2024}

\usepackage{amsmath}
\usepackage{amssymb}
\usepackage{mathtools}
\usepackage{amsthm}

\usepackage{multirow}
\usepackage{pifont}
\usepackage{multicol}
\usepackage{bbm}

\usepackage[capitalize,noabbrev]{cleveref}

\theoremstyle{plain}
\newtheorem{theorem}{Theorem}[section]

\newtheorem{lemma}[theorem]{Lemma}

\theoremstyle{definition}
\newtheorem{definition}[theorem]{Definition}

\theoremstyle{remark}

\usepackage[textsize=tiny]{todonotes}

\icmltitlerunning{Cluster-Aware Similarity Diffusion for Instance Retrieval}

\begin{document}

\twocolumn[
\icmltitle{Cluster-Aware Similarity Diffusion for Instance Retrieval}



\icmlsetsymbol{equal}{*}

\begin{icmlauthorlist}
\icmlauthor{Jifei Luo}{ustc}
\icmlauthor{Hantao Yao}{casia,ucas}
\icmlauthor{Changsheng Xu}{casia,ucas}
\end{icmlauthorlist}

\icmlaffiliation{ustc}{University of Science and Technology of China, Hefei, China}
\icmlaffiliation{casia}{Institute of Automation, Chinese Academy of Sciences, Beijing, China}
\icmlaffiliation{ucas}{University of Chinese Academy of Sciences}

\icmlcorrespondingauthor{Hantao Yao}{hantao.yao@nlpr.ia.ac.cn}

\icmlkeywords{Machine Learning, ICML}

\vskip 0.3in
]



\printAffiliationsAndNotice{}  

\begin{abstract}
Diffusion-based re-ranking is a common method used for retrieving instances by performing similarity propagation in a nearest neighbor graph.
However, existing techniques that construct the affinity graph based on pairwise instances can lead to the propagation of misinformation from outliers and other manifolds, resulting in inaccurate results.
To overcome this issue, we propose a novel Cluster-Aware Similarity (CAS) diffusion for instance retrieval.
The primary concept of CAS is to conduct similarity diffusion within local clusters, which can reduce the influence from other manifolds explicitly.
To obtain a symmetrical and smooth similarity matrix, our Bidirectional Similarity Diffusion strategy introduces an inverse constraint term to the optimization objective of local cluster diffusion.
Additionally, we have optimized a Neighbor-guided Similarity Smoothing approach to ensure similarity consistency among the local neighbors of each instance.
Evaluations in instance retrieval and object re-identification validate the effectiveness of the proposed CAS, our \href{https://github.com/jifeiluo/reranking}{code} is publicly available.
\end{abstract}

\section{Introduction}
\label{section: introduction}

Instance retrieval aims to search through a large-scale database to identify images that share similar content with a given image.
Recently, traditional descriptors \cite{ijcv2004_sift, tpami2012_fisher_vector} are gradually being supplanted by global descriptors extracted with deep neural networks \cite{tpami2019_rtc, iccv2021_dolg, cvpr2023_senet} for instance retrieval.
Due to factors such as illumination, occlusion, and variations in viewpoint among images, the global descriptors may not yield optimal retrieval results.
Therefore, refining the initial retrieval results to enhance the retrieval performance becomes a pivotal task, also known as re-ranking.
Among re-ranking methods, Query Expansion (QE) \cite{iccv2007_aqe, cvpr2012_dqe, ijcv2017_end2end, eccv2020_lattqe} is a representative approach that involves the weighted summation of features from top-ranked images to generate enhanced features for a secondary round of retrieval.
However, QE is a primitive operation that cannot explore the data manifold structure.

To model the intricacies of the underlying data manifold, diffusion-based re-ranking methods \cite{cvpr2013_diffusion_processes, cvpr2017_dfs, cvpr2018_fsr, aaai2019_eir, cvpr2019_egt, tmm2019_Deep_Feature_Aggregation, ICCV2017_Ensemble_Diffusion, cvpr2019_ued} have been proposed to perform similarity propagation within the nearest neighbor graph.
The concept of ranking data concerning the intrinsic global manifold structure in a diffusion manner was initially introduced by \cite{nips2003_ranking_on_data_manifolds}. 
Various methods \nocite{tpami2008_visualrank}\cite{cvpr2009_lcdp, cvpr2013_diffusion_processes, tmm2016_Diffusion, tmm2023_Graph_Convolution} propose to iteratively propagate the similarity among pairwise neighbors in the graph during each iteration.
To capture the complex relationships between instances, some studies \cite{tpami2013_tpg, tpami2015_qsr, tpami2019_rdp} expand the diffusion theory to a hypergraph, thereby considering higher-order information.
However, since these graphs and hypergraphs are constructed based on pairwise instances, the similarity propagation is prone to mistakenly spread information from outliers and other manifolds throughout the entire graph, leading to suboptimal performance.
Therefore, mitigating the influence of samples from other manifolds during similarity propagation is crucial for enhancing the robustness of diffusion-based re-ranking.

In diffusion-based re-ranking methods, the connections of outliers or other manifolds to individual instances can reduce the reliability of ranking during the process of similarity propagation.
Therefore, the intuitive idea is to use the neighborhood cluster of each individual instance to guide the process of diffusion.
Different from existing methods that propagate the similarity among the entire graph, we aim to constrain the similarity propagation among the local clusters consisting of the individual instance and its neighbors.
Most of the undesired instances can be explicitly filtered out by the approximated local cluster, allowing us to benefit from reducing the impact of other manifolds.
Based on the assumption that samples belonging to the same cluster should be similar, an individual instance and its neighbor samples should have a consistent similarity with other instances.
Compared to traditional diffusion-based methods that only use instance-to-instance similarity, leveraging local neighbors to optimize the similarity matrix can be beneficial in reducing the impact of anomalous instances within the local cluster, potentially leading to better retrieval results.
Moreover, inspired by QE, the local neighbors can be used to further enhance the representation of the similarity matrix.
In conclusion, the cluster can be utilized to guide the construction of a similarity matrix, thereby enhancing the re-ranking performance of the diffusion-based model. 

By considering the above issues, we propose a novel Cluster-Aware Similarity (CAS) diffusion for instance retrieval, consisting of Bidirectional Similarity Diffusion (BSD) and Neighbor-guided Similarity Smooth (NSS).
Given the initial similarities of all samples, Bidirectional Similarity Diffusion (BSD) first generates a local cluster for each instance.
After that, by introducing an inverse similarity constraint term, Bidirectional Similarity Diffusion is optimized to obtain a symmetrical and smooth similarity matrix within the local cluster diffusion process.
To further suppress the influence from anomalous instances and other manifolds, NSS ensures the similarity consistency among neighbors by leveraging the average similarity between instances and local neighbors to refine the matrix.
Moreover, the local neighbors can also be used to enhance the representation of the similarity matrix. 
Finally, to better fit the global retrieval task, we propagate the similarity matrix within the graph, which is then applied for global instance retrieval.

Evaluations in instance retrieval and re-identification validate the effectiveness of the proposed Cluster-Aware Similarity diffusion.
For example, CAS achieves an mAP of 80.7\%/64.8\% on the medium and hard tasks of \emph{R}Oxf, and 91.0\%/80.7\% on \emph{R}Par, demonstrating its superior performance.

\section{Related Work}
\label{section: related work}

\textbf{Instance Retrieval.} 
In instance retrieval tasks, the objective is to identify images within a database that either depict the same object as the query image or portray a scene similar to it.
With the rapid advancement of deep learning, global features extracted by using neural networks \cite{iccv2021_dolg, cvpr2022_cvnet, cvpr2023_senet} have surpassed traditional local descriptors \cite{ijcv2004_sift, tpami2012_fisher_vector} in both speed and accuracy, thereby dominating the field of image retrieval tasks. 
In this paper, the search results obtained by global features are referred to as initial ranking.
Person re-identification (ReID) is a special kind of instance retrieval, it aims at matching the same identities captured by different camera views.
In recent years, there has been a significant improvement in the performance of ReID \cite{tpami2022_agw, mm2018_mgn, nips2020_spcl, aaai2023_clip_reid, cvprw2019_bot, cvpr2022_ise_reid, tpami2022_osnet, iccv2019_abd_net, cvpr2023_adasp_reid, iccv2021_trans_reid}.

\textbf{Re-ranking.} 
Re-ranking enhances the overall retrieval performance by refining the initial ranking list through a second round of retrieval or optimization techniques, which can be divided into \textit{Query Expansion}, \textit{Diffusion-based Method}, \textit{Context-based Method}, and \textit{Learning-based Method}.

\textit{Query Expansion.} 
Based on the observation that top-ranked images have the potential capability to enhance ranking performance, Query Expansion (QE) seeks to aggregate the neighboring features to construct a stronger and more descriptive query.
AQE \cite{iccv2007_aqe} directly average the top-$k$ returned image features, while AQEwD \cite{ijcv2017_end2end} and $\alpha$QE \cite{tpami2019_rtc} proposes a monotonically decreasing weights to mitigate the impact of later images. 
To make use of the bottom-ranked negative samples, DQE \cite{cvpr2012_dqe} trains a linear SVM to resort the gallery images. 
And the recently proposed SG \cite{iccv2023_super_global} further refines the QE strategy, leading to enhanced retrieval efficiency.

\textit{Diffusion-based Method.} 
\nocite{icml2022_diffusion_on_hypergraphs}
Diffusion-based method is a powerful re-ranking technique that can leverage the intrinsic manifold structure of data, its principles and applications have been studied \cite{nips2003_ranking_on_data_manifolds, cvpr2009_lcdp, cvpr2013_diffusion_processes} for years.
As the most representative works, DFS \cite{cvpr2017_dfs} and FSR \cite{cvpr2018_fsr} have achieved excellent results in instance retrieval tasks.
In order to effectively aggregate higher-order information, 
\cite{tpami2019_rdp, tpami2013_tpg, tpami2015_qsr} construct a hypergraph for diffusion. 
Besides, Fusion with Diffusion \cite{nips2012_fusion_with_diffusion} manages to integrate the manifold information from distinct affinity graphs, later works \cite{tip2019_red, cvpr2019_ued} further assign automatically learned weights and generalize the procedure into a unified framework.
Additionally, EGT \cite{cvpr2019_egt} adjusts the diffusion process through a two-stage approach, resulting in improved effectiveness.

\textit{Context-based Method.} 
The relevant contextual information contained by $k$-nearest neighbors holds the potential to improve retrieval performance significantly. 
CDM \cite{cvpr2007_cdm} iteratively modifies the neighborhood structure, kNN \cite{cvpr2012_knn_reranking} recalculates the distance measure based on the rank lists of $k$-nearest neighbors, while ECN \cite{cvpr2018_ecn} introduces the concept of expanded cross neighborhood for aggregating a new rank list.
Moreover, SCA \cite{tip2016_sca}, $k$-reciprocal \cite{cvpr2017_k_reciprocal} and STML \cite{cvpr2022_stml} encode each instance into a contextual affinity feature space, where similar images exhibit higher consistency.
ConAff \cite{arxiv2020_gnn_reranking, nips2023_connr} further enhances feature representation by propagating information within the affinity graph.

\textit{Learning-based Method.}
Recently, self-attention mechanism and graph neural network (GNN) have been introduced into the field of visual re-ranking. 
LAttQE \cite{eccv2020_lattqe} leverages a transformer encoder to learn affinity weights for query expansion, while CSA \cite{nips2021_csa}  represents each instance as an affinity feature and then aggregates the contextual information with a self-attention mechanism.
By adapting the properties of graph neural network \cite{iclr2017_gcn, iclr2018_predict}, GSS \cite{nips2019_gss} and SSR \cite{nips2021_ssr} can obtain a more representative descriptor by optimizing the graph model.
\nocite{icml2021_clip, science_lle, science_isomap}

\section{Background}
\label{section: background}

Given a query image $x_q$, the instance retrieval aims to sort the gallery image set $\mathcal{X}_g$ in ascending order based on a distance metric. We say that the higher-ordered image has a higher probability of having the same label as the query instance. 
Formally, we define the query set as $\mathcal{X}_q=\{x_1,x_2,...,x_{n_q}\}$, and the gallery set as $\mathcal{X}_g=\{x_1,x_2,...,x_{n_g}\}$, where $n_q$ and $n_g$ are the number of images in query and gallery sets, respectively.
The whole image set can be represented as $\mathcal{X}=\{x_1, x_2,...,x_n\}$, in which $n=n_q+n_g$. 
For each image in $\mathcal{X}$, a $d$-dimension deep feature is extracted for retrieval. 
Defining the feature of $x_i$ as $\boldsymbol{f}_i$, the pairwise distance between images $x_i$ and $x_j$ can be calculated using the Euclidean distance between their features as follows:
\begin{equation}
\label{eq: euclidean distance}
    d(i, j) = \Vert\boldsymbol{f}_i-\boldsymbol{f}_j\Vert_2.
\end{equation}

After computing the distance among each pair of images, we can obtain the initial distance matrix $\boldsymbol{M}\in\mathbb{R}^{n\times n}$, where $\boldsymbol{M}_{ij}=d(i,j)$.
Each row $\boldsymbol{M}_i$ represents the distance between the image $x_i$ and image set $\mathcal{X}$, which is used to sort the gallery images with respect to the query image $x_i$.

Based on the prior knowledge that similar images are lying on a low-dimensional manifold structure contained by the distance matrix $\boldsymbol{M}$, we can map $\boldsymbol{M}$ into a new space to ensure that instances with high rankings are not only close in Euclidean space but also exhibit higher proximity in the manifold space.
Moreover, the manifold structure can be approximated by utilizing the feature set $\mathcal{X}$ to construct an $k$-nearest neighbor graph $\mathcal{G}=\{\mathcal{V},\mathcal{E}\}$. 
The vertices set $\mathcal{V}=\{{v}_1, {v}_2, \dots, {v}_n\}$ represents a corresponding feature in $\mathcal{X}$, while $\mathcal{E}$ are the edges between two vertices.
The edge weights are represented as:
\begin{equation}
    \boldsymbol{W}_{ij}=\mathbbm{1}_{ij}\exp{(-d^2(i,j)/\sigma^2)},
\end{equation}
where $\mathbbm{1}$ is an indicator, denote the $k$-nearest neighbors belongs to $x_i$ as $\mathcal{N}(i,k)$, then $\mathbbm{1}_{ij}=1, \text{if\ }j\in\mathcal{N}(i,k)$.

Based on the $k$-nearest neighbor graph inferred by affinity weight matrix $\boldsymbol{W}$, 
the diffusion-based methods \cite{nips2003_ranking_on_data_manifolds, cvpr2017_dfs, aaai2017_rdp} propagates the information through the graph $\mathcal{G}$ for inferring a similarity matrix $\boldsymbol{F}$.
Since the propagation progress is over the whole graph, the obtained similarity matrix $\boldsymbol{F}$ is susceptible to the negative impacts from outlier and nearby manifolds, which yields bad performance for instance retrieval.

\begin{figure}
    \centering
    \includegraphics[width=0.8\linewidth]{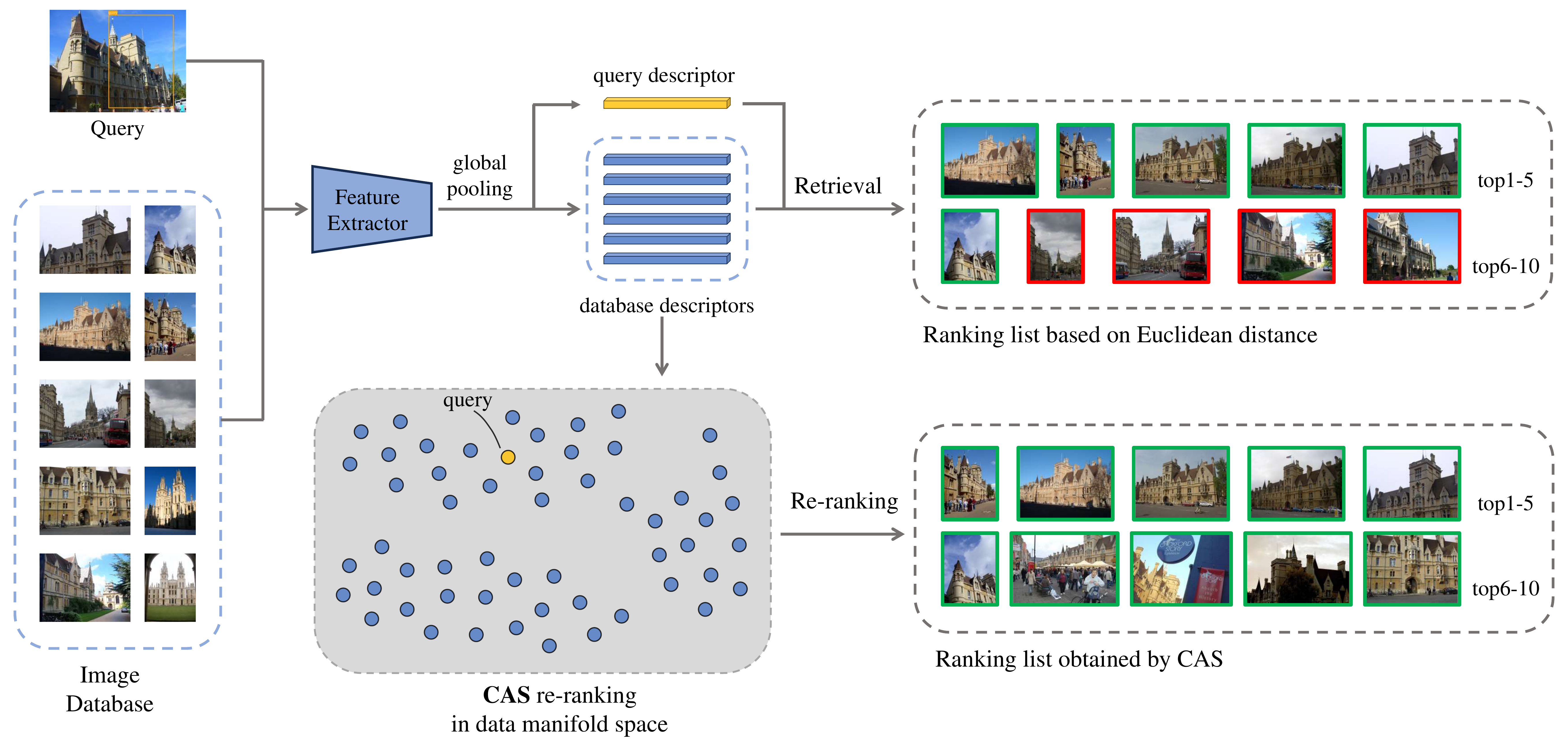}
    \vspace{-0.5em}
    \caption{The workflow of Cluster-Aware Similarity Diffusion.}
    \label{fig: workflow}
    \vskip -0.1in
    \vspace{-0.5em}
\end{figure}

\section{Cluster-Aware Similarity Diffusion}
Different from existing methods that propagate the similarity through the entire graph, we constrain the diffusion process to the local graph consisting of the individual instance and its neighbors, which can help mitigate the affection from other manifolds.
Therefore, we propose a novel Cluster-Aware Similarity (CAS) diffusion to optimize the similarity matrix $\boldsymbol{F}$.
Especially, an inverse similarity term is introduced in the Bidirectional Similarity Diffusion (BSD) for obtaining a symmetrical and smooth similarity matrix $\boldsymbol{F}$ with the constraint from the approximated cluster of each instance.
Subsequently, the Neighbor-guided Similarity Smooth (NSS) is further proposed to refine the similarity matrix $\boldsymbol{F}$ by ensuring similarity consistency among the local neighbors of each instance.
The workflow is shown in \cref{fig: workflow}, in the following, we will give a detailed introduction to the Bidirectional Similarity Diffusion and Neighbor-guided Similarity Smooth.

\subsection{Bidirectional Similarity Diffusion}
\label{Bidirectional Similarity Diffusion}
Bidirectional Similarity Diffusion seeks to perform similarity propagation on the local graph instead of the entire graph, such that the dissemination of misleading information from other noisy samples can be diminished.
For each instance, the local graph can be treated as a local cluster $\mathcal{C}$ that includes the instance itself and its similar neighbors, \emph{e.g.}, $\mathcal{C}[i]$ represents the local cluster of the image ${x}_i$.
The smoothed similarity matrix $\boldsymbol{F}$ can be optimized by solving the following objective function:
\begin{equation}
\label{eq: obj}
    \begin{aligned}
    \min_{\boldsymbol{F}}\quad  &\frac{1}{4}\sum_{k=1}^{n}\sum_{i,j\in\mathcal{C}[k]} \boldsymbol{W}_{ij}\Big(\frac{\boldsymbol{F}_{ki}}{\sqrt{\boldsymbol{D}_{ii}}} - \frac{\boldsymbol{F}_{kj}}{\sqrt{\boldsymbol{D}_{jj}}}\Big)^2 \\&+ \boldsymbol{W}_{ij}\Big(\frac{\boldsymbol{F}_{ik}}{\sqrt{\boldsymbol{D}_{ii}}} - \frac{\boldsymbol{F}_{jk}}{\sqrt{\boldsymbol{D}_{jj}}}\Big)^2 + \mu\Vert\boldsymbol{F}-\boldsymbol{E}\Vert^2_F, \\
    \mbox{s.t.}\quad &\boldsymbol{F}_{ij}=0,\quad j\notin \mathcal{C}[i]
    \end{aligned}
\end{equation}
where $\boldsymbol{W}$ is a symmetric matrix representing the adjacent weights of the nearest neighbor graph.
$\boldsymbol{D}$ is the diagonal matrix with its $i$-th diagonal element equal to the summation of the $i$-th row in $\boldsymbol{W}$. 
Matrix $\boldsymbol{E}$ in the regularization term is positive and semi-definite, which can prevent $\boldsymbol{F}$ from excessively smooth.
While $\mu>0$ represents the constraint weight.
For a triplet of instances ${x}_k$, ${x}_i$ and ${x}_j$ within the local cluster, the edge weight $\boldsymbol{W}_{ij}$ is used to regularize the similarity between $\boldsymbol{F}_{ki}$ and $\boldsymbol{F}_{kj}$. 
Meanwhile, the pair of similarities $\boldsymbol{F}_{ik}$ and $\boldsymbol{F}_{jk}$ is also taken into account, such that the smooth strategy is bidirectional, ensuring the symmetric of the obtained similarity matrix $\boldsymbol{F}$.

In addition, we incorporate $k$-reciprocal neighbors \cite{cvpr2011_hello_neighbor, cvpr2017_k_reciprocal} to construct the local cluster $\mathcal{C}$, which can be viewed as a stricter case of $k$-nearest neighbors by considering reverse information.
This strategy is effective in diminishing the presence of noises within the neighbor set, thereby minimizing their influence during the cluster-aware smoothing process.
The definition of $k$-reciprocal neighbors is formulated as:
\begin{equation}
    \mathcal{R}(i,k) = \{j|(j\in\mathcal{N}(i,k)) \wedge (i\in\mathcal{N}(j,k))  \},
\end{equation}
To avoid ambiguity and reduce the number of parameters, we set the value of $k$ used to construct the local cluster as $k_1$. 
Additionally, the $k$-reciprocal neighbors of an image $x_i$ can be expanded from $\mathcal{R}(i,k_1)$ to a larger set $\mathcal{R}^*(i,k_1)$ when the number of elements within $\mathcal{R}(i,k_1)$ reaches a certain threshold.

However, it is difficult to directly optimize Eq.~\eqref{eq: obj}.
We thus simplify the optimization problem as two sub-problems by relaxing the smoothing area and the constraint conditions.
Firstly, we aim to solve the optimization problem without considering the cluster constraint with Eq.~\eqref{eq: obj2},
\begin{equation}
\label{eq: obj2}
    \begin{aligned}
    \min_{\boldsymbol{F}}\  &\frac{1}{4}\sum_{k=1}^{n}\sum_{i,j=1}^{n} \boldsymbol{W}_{ij}\Big(\frac{\boldsymbol{F}_{ki}}{\sqrt{\boldsymbol{D}_{ii}}} - \frac{\boldsymbol{F}_{kj}}{\sqrt{\boldsymbol{D}_{jj}}}\Big)^2 \\&+ \boldsymbol{W}_{ij}\Big(\frac{\boldsymbol{F}_{ik}}{\sqrt{\boldsymbol{D}_{ii}}} - \frac{\boldsymbol{F}_{jk}}{\sqrt{\boldsymbol{D}_{jj}}}\Big)^2 + \mu\Vert\boldsymbol{F}-\boldsymbol{E}\vert^2_F.
    \end{aligned}
\end{equation}
Then, we only update the similarity terms for samples belonging to the corresponding cluster $\mathcal{C}$, which serves as an effective approximation for performing similarity propagation within local clusters.
To facilitate the solution of the optimization problem in Eq.~\eqref{eq: obj2}, we leverage certain graph theory tools to reformulate the objective function as follow\footnote{The proof is shown in \cref{section: appendix bidirectional similarity diffusion process}}:
\begin{equation}
    \label{eq: method approximate objective}
    \begin{aligned}
    J = vec(\boldsymbol{F})^T\big(\mathbb{I}-\mathbb{\Bar{S}}\big)vec(\boldsymbol{F})+ \mu\Vert vec(\boldsymbol{F}-\boldsymbol{E})\Vert_2^2,
    \end{aligned}
\end{equation}
where $\mathbb{I}\in\mathbb{R}^{n^2\times n^2}$ is a diagonal matrix.
The normalized matrix of $\boldsymbol{W}$ is represented as $\boldsymbol{S}=\boldsymbol{D}^{-1/2}\boldsymbol{W}\boldsymbol{D}^{1/2}$, and
$\mathbb{\Bar{S}}\in \mathbb{R}^{n^2\times n^2}$ is a mean Kronecker product calculated by $\mathbb{\Bar{S}}=(\boldsymbol{S}\otimes \boldsymbol{I} + \boldsymbol{I}\otimes{\boldsymbol{S}})/2$.
The term $vec^{-1}(\cdot)$ denotes the inverse function of $vec(\cdot)$, which is a vectorization or flattening operation.

The optimization problem in Eq.~\eqref{eq: method approximate objective} is convex with the condition that its Hessian matrix is positive definite.
The Hessian matrix $\boldsymbol{H}$ of the objective function is: 
\begin{equation}
    \label{eq: method hessian matrix}
    \begin{aligned}
        \boldsymbol{H} = \nabla^2_{vec(F)}J = 2(\mathbb{I}-\Bar{\mathbb{S}}) + 2\mu\mathbb{I}.
    \end{aligned}
\end{equation}
\begin{lemma}\label{lemma: method radius constrain}
    Let $\boldsymbol{A}\in\mathbb{R}^{n\times n}$, the spectral radius of $\boldsymbol{A}$ is denoted as $\rho(\boldsymbol{A})=\max\{|\lambda|,\lambda\in\sigma(\boldsymbol{A})\}$, where $\sigma(\boldsymbol{A})$ is the spectrum of $\boldsymbol{A}$ that represents the set of all the eigenvalues. 
    Let $\Vert\cdot\Vert$ be a matrix norm on $\mathbb{R}^{n\times n}$, given a square matrix $\boldsymbol{A}\in\mathbb{R}^{n\times n}$, $\lambda$ is an arbitrary eigenvalue of $\boldsymbol{A}$, then we have $|\lambda|\leq\rho(\boldsymbol{A})\leq\Vert\boldsymbol{A}\Vert$.
\end{lemma}
\begin{lemma}\label{lemma: method kronecker eigenvalue}
    Let $\boldsymbol{A}\in\mathbb{R}^{m\times m}$, $\boldsymbol{B}\in\mathbb{R}^{n\times n}$, denote $\{\lambda_i,\boldsymbol{x}_i\}_{i=1}^{m}$ and $\{\mu_i,\boldsymbol{y}_i\}_{i=1}^{n}$ as the eigen-pairs of $\boldsymbol{A}$ and $\boldsymbol{B}$ respectively. The set of $mn$ eigenpairs of $\boldsymbol{A}\otimes \boldsymbol{B}$ is given by
    \begin{equation}
        \{\lambda_i\mu_j, \boldsymbol{x}_i\otimes \boldsymbol{y}_j\}_{i=1,\dots,m,\ j=1,\dots n},
    \end{equation}
\end{lemma}
\emph{Proof of the positive definiteness of the Hessian matrix $\boldsymbol{H}$}:
Consider the matrix $\boldsymbol{D}^{-1}\boldsymbol{W}$, since $\boldsymbol{D}$ is a diagonal matrix with its $i$-th element the sum of the corresponding $i$-th row of matrix $\boldsymbol{W}$, we can obtain that $\Vert\boldsymbol{D}^{-1}\boldsymbol{W}\Vert_{\infty}=1$. 
According to \cref{lemma: method radius constrain}, all the eigenvalues are no larger than 1, \emph{i.e.}, $\rho(\boldsymbol{D}^{-1}\boldsymbol{W})\leq1$.
As for the normalized matrix $\boldsymbol{S}=\boldsymbol{D}^{-1/2}\boldsymbol{W}\boldsymbol{D}^{-1/2}$ we are concerned about, we can rewrite it as $\boldsymbol{D}^{1/2}\boldsymbol{D}^{-1}\boldsymbol{W}\boldsymbol{D}^{-1/2}$, which implies that it is similar to the matrix $\boldsymbol{D}^{-1}\boldsymbol{W}$.
Since two similar matrices share the same eigenvalues, it can be inferred that $\rho(\boldsymbol{S})\leq1$.
By applying \cref{lemma: method kronecker eigenvalue}, the spectral radius of the mean Kronecker product $\mathbb{\Bar{S}}=\boldsymbol{S}\otimes \boldsymbol{I} + \boldsymbol{I}\otimes \boldsymbol{S}$ is not exceeding $1$, \emph{i.e.}, $\rho(\mathbb{\Bar{S}})\leq1$.
Since the constraint weight $\mu>0$, we can conclude that the Hessian matrix of Eq.~\eqref{eq: method hessian matrix} is positive definite.

Therefore, the optimization problem in Eq.~\eqref{eq: method approximate objective} is convex, and we can obtain the optimal solution by taking the partial derivative of $vec(\boldsymbol{F})$, that is:
\begin{equation}
    \label{eq: method partial derivative}
    \nabla_{vec(\boldsymbol{F})}J = 2(\mathbb{I}-\Bar{\mathbb{S}})vec(\boldsymbol{F}) + 2\mu(vec(\boldsymbol{F}-\boldsymbol{E})).
\end{equation}
By setting the value of Eq.~\eqref{eq: method partial derivative} to zero, the closed form solution $\boldsymbol{F}^*$ can be obtained by Eq.~\eqref{eq: method closed form solution},
\begin{equation}
    \label{eq: method closed form solution}
    \boldsymbol{F}^* = (1-\alpha)vec^{-1}\big((\mathbb{I}-\alpha\mathbb{\Bar{S}})^{-1}vec(\boldsymbol{E})\big).
\end{equation}
Here we substitute the hyper-parameter $\mu$ with $\alpha=\frac{1}{1+\mu}$ to simplify the expression, and a more general case is discussed in 
\cref{section: appendix bidirectional similarity diffusion process} when $\boldsymbol{W}$ is not symmetric. 
Moreover, the optimum solution to the slack optimization can also be derived by solving the following Lyapunov equation:
\begin{equation}
    (\boldsymbol{I}-\alpha\boldsymbol{S})\boldsymbol{F} + \boldsymbol{F}(\boldsymbol{I}-\alpha\boldsymbol{S}) = 2(1-\alpha)\boldsymbol{E}.
\end{equation}

By taking the graph $\mathcal{G}=\{\mathcal{V},\mathcal{E}\}$ as a progressively stable linear system defined by $\dot{z}=(\alpha\boldsymbol{S}-\boldsymbol{I})z$, with $z$ representing the status of each vertex, a Lyapunov energy induced by $\boldsymbol{F}$ of the linear system can be solved, where $\boldsymbol{F}$ contains the manifold distance information instinctively.
Since it is infeasible to directly solve the closed-form of ~Eq.~\eqref{eq: method closed form solution} in terms of time complexity, we can adopt an iterative approach\footnote{\cref{alg: efficient iterative solution} introduces a more efficient iteration with a higher convergence rate.} to gradually converge towards the solution, making the computational process more feasible and efficient,
\begin{equation}
    \boldsymbol{\boldsymbol{F}}^{(t+1)} = \frac12\alpha \big(\boldsymbol{F}^{(t)}\boldsymbol{S}^\top + \boldsymbol{S}\boldsymbol{F}^{(t)}\big) + (1-\alpha)\boldsymbol{E}.
\end{equation}
As proved in \cref{appendix: basic iterative solution}, $\boldsymbol{F}^{(t)}$ will converge to the closed-form solution, allowing us to numerically solve the optimization problem. 
After obtaining the relaxed similarity matrix $\boldsymbol{F}^*$, we selectively preserve similarity between each point and its local cluster $\mathcal{C}$, following the constraint conditions.
Taking an instance $x_i$ for example, only the terms corresponding to the local cluster $\mathcal{C}[i]$ are assigned with similarity solved by Eq.\eqref{eq: method closed form solution} to obtain the cluster-aware smoothed similarity matrix $\boldsymbol{F}$.
Subsequently, an $l_1$-normalization operation is applied to each row to obtain the final smoothed similarity matrix.

\begin{figure}
    \centering
    \includegraphics[width=0.8\linewidth]{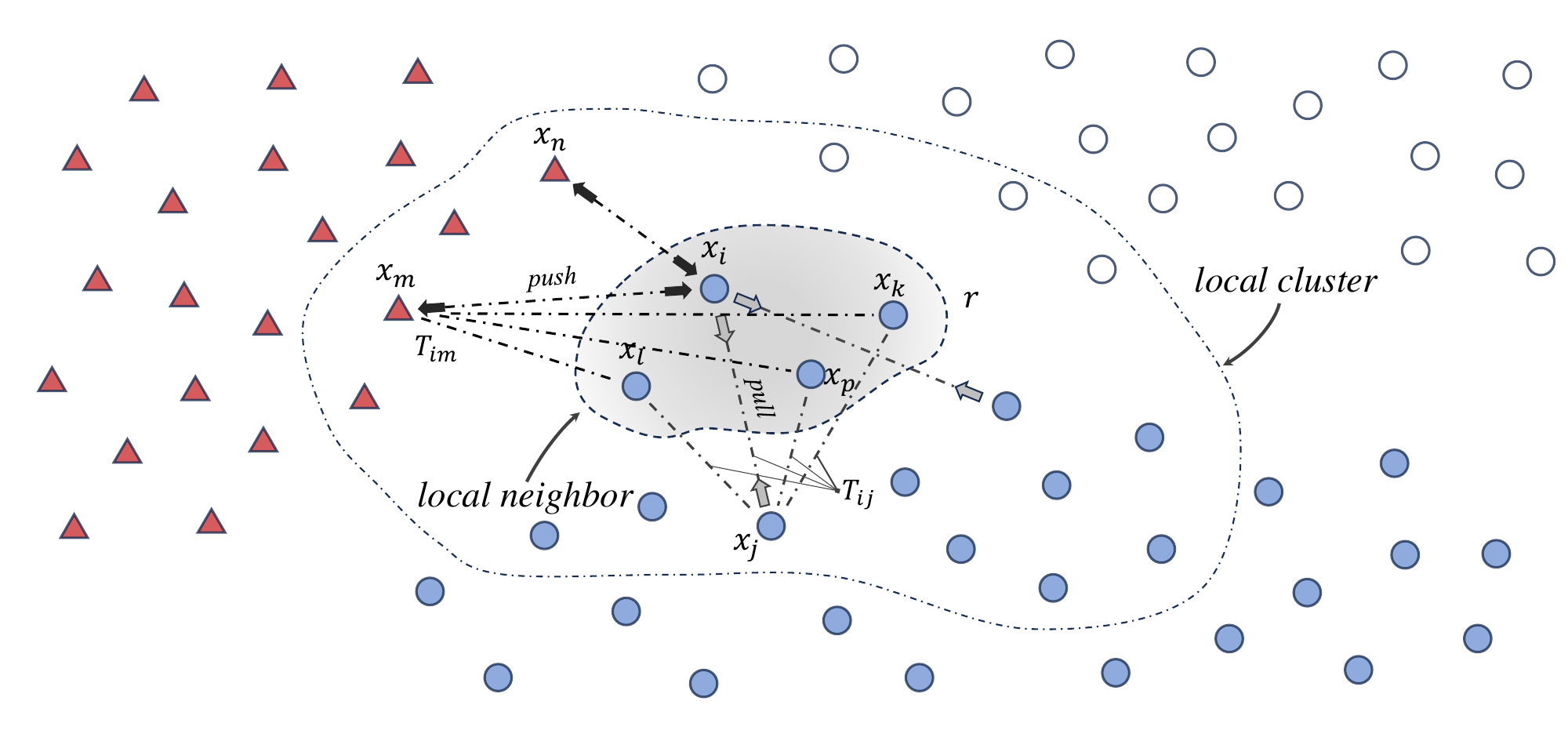}
    \caption{Explanation of the Neighbor-guided Similarity Smooth strategy. For an instance $x_i$, the average similarity $\boldsymbol{T}_{ij}$ from $x_j$ to its local neighbor is used to constrain the similarity $\boldsymbol{F}_{ij}$. Which can help suppress the influence of instances from different manifolds.}
    \label{fig:enhancement}
    \vspace{-0.5em}
\end{figure}

\subsection{Neighbor-guided Similarity Smooth}
\label{Neighbor-guided Similarity Smooth}
Local neighbors can provide a more precise estimation of inter-class relationships compared with individual points. 
Therefore, we aim to leverage the heuristic information from the local neighbor set $\xi$ to further enhance the smoothness of the similarity matrix by constraining the similarity consistency from instance to local neighbors, and then adjusting it to fit the global retrieval task.

Based on the above issue, the Neighbor-guided Similarity Smooth (NSS) is proposed to initially refine the obtained similarity matrix $\boldsymbol{F}$ into a neighbor consistent matrix $\boldsymbol{\hat{F}}$.
Subsequently, by leveraging $\boldsymbol{\hat{F}}$, an enhanced representation $\boldsymbol{\widetilde{F}}$ is obtained, finally, $\boldsymbol{\widetilde{F}}$ is extended to the entire graph to generate $\boldsymbol{F}'$, thereby accommodating global retrieval tasks.
As shown in \cref{fig:enhancement}, for an instance ${x}_i$, its local neighbors $\xi[i]$ obtained by $\mathcal{R}(i,k_2)$ are indicated by shades, \emph{i.e.}, $k_2<k_1$.
The objective is to jointly optimize the similarity from ${x}_i$ to other images by constraining the local neighbor consistency.
Specifically, the optimized similarity between ${x}_i$ and ${x}_j$ should be consistent with the factor $\boldsymbol{T}_{ij}/r$, which is a normalized value that measures the proximity from an image ${x}_j$ to the neighbors of ${x}_i$.  
In this case, we define $\boldsymbol{T}_{ij}$ as the average similarities from image $x_j$ to the local neighbors for image $x_i$.
Additionally, the average similarity between pairwise instances within the local neighborhood $\xi$ is defined as $r$ to represent the reliability of local neighbors.
Since a larger $r$ implies a higher likelihood that neighbors belong to the same cluster, here we use $r^2$ to constrain the adjustment.
By considering the above issue, the optimization objective can be formulated as:
\begin{equation}
    \label{eq: method similarity smooth}
    \begin{aligned}
        \mbox{minimize} \quad& \frac{1}{2}r^2\Vert\boldsymbol{\hat{F}}_i-\frac{\boldsymbol{T}_i}{r}\circ\boldsymbol{F}_i\Vert^2_2 + \beta\Vert\boldsymbol{\hat{F}}_i-\boldsymbol{F}_i\Vert^2_2, \\
        \mbox{subject to}\quad & \boldsymbol{\hat{F}}_{ij}\geq0,\quad j\in\mathcal{C}[i] \\
        & \boldsymbol{\hat{F}}_{ij}=0, \quad j=1,2\dots,\mathcal{|X|}/\mathcal{C}[i]\\
                    & \Vert\boldsymbol{\hat{F}}_i\Vert_1=\Vert\boldsymbol{F}_i\Vert_1
    \end{aligned}
\end{equation}
where symbol $\circ$ denotes that two vectors are multiplied by the corresponding element. $\mathcal{C}$ is the local cluster set that we used to constrain the similarity connections as introduced in \cref{Bidirectional Similarity Diffusion}.
The regularization term weighted by $\beta$ can constrain the value of $\boldsymbol{\hat{F}}_i$ from deviating too much from the initial value $\boldsymbol{F}$, we set $\beta$ with a low value to achieve better numerical stability.
Moreover, here we truncate the vector $\boldsymbol{T}_i$ to ensure that all its values are no larger than $r$.

The Lagrange function $\mathcal{L}(\boldsymbol{\hat{F}}_i,\boldsymbol{\lambda},\nu)$ corresponding to the primal constrained optimization problem Eq.~\eqref{eq: method similarity smooth} is formulated as below,
\begin{equation}
    \begin{aligned}
        \mathcal{L}(\boldsymbol{\hat{F}}_i,\boldsymbol{\lambda},\nu)=&\frac12r^2\Vert\boldsymbol{\hat{F}}_i-\frac{\boldsymbol{T}_i}{r}\circ\boldsymbol{F}_i\Vert^2_2 + \beta\Vert\boldsymbol{\hat{F}}_i-\boldsymbol{F}_i\Vert^2_2 \\
        -&\sum_{j=1}^{n}\boldsymbol{\lambda}_j\boldsymbol{\hat{F}}_{ij} + \nu(\sum_{j=1}^n \boldsymbol{\hat{F}}_{ij} - \sum_{j=1}^n \boldsymbol{F}_{ij}).
    \end{aligned} 
\end{equation}

\begin{table*}[t]
\caption{Evaluation of the Performance on \emph{R}Oxf, \emph{R}Par, \emph{R}Oxf+1M, \emph{R}Par+1M. Using R-GeM \cite{tpami2019_rtc} as the baseline.}
\label{tab: rgem}
\vskip 0.1in
\centering
\resizebox{0.92\textwidth}{!}{
\begin{tabular}{lcccccccc}
\toprule[1pt]
\multicolumn{1}{c}{\multirow{2.4}*{Method}} &  \multicolumn{4}{c}{Medium} & \multicolumn{4}{c}{Hard} \\
\cmidrule(lr){2-5}
\cmidrule(lr){6-9}
~ & \emph{R}\textbf{Oxf} & \emph{R}\textbf{Oxf}+\textbf{1M} & \emph{R}\textbf{Par} & \emph{R}\textbf{Par}+\textbf{1M} & \emph{R}\textbf{Oxf} & \emph{R}\textbf{Oxf}+\textbf{1M} & \emph{R}\textbf{Par} & \emph{R}\textbf{Par}+\textbf{1M} \\
\midrule
R-GeM \cite{tpami2019_rtc} & 67.3 & 49.5 & 80.6 & 57.4 & 44.2 & 25.7 & 61.5 & 29.8  \\
\midrule
AQE \cite{iccv2007_aqe} & 72.3 & 56.7 & 82.7 & 61.7 & 48.9 & 30.0 & 65.0 & 35.9  \\
$\alpha$QE \cite{tpami2019_rtc} & 69.7 & 53.1 & 86.5 & 65.3 & 44.8 & 26.5 & 71.0 &  40.2 \\
DQE \cite{cvpr2012_dqe} & 70.3 & 56.7 & 85.9 & 66.9 & 45.9 & 30.8 & 69.9 & 43.2  \\
AQEwD \cite{ijcv2017_end2end} & 72.2 & 56.6 & 83.2 & 62.5 & 48.8 & 29.8 & 65.8 & 36.6  \\
LAttQE \cite{eccv2020_lattqe} & 73.4 & 58.3 & 86.3 & 67.3 & 49.6 & 31.0 & 70.6 &  42.4  \\
\midrule
ADBA+AQE & 72.9 & 52.4 & 84.3 & 59.6 & 53.5 & 25.9 & 68.1 & 30.4  \\
$\alpha$DBA+$\alpha$QE & 71.2 & 55.1 & 87.5 & 68.4 & 50.4 & 31.7 & 73.7 & 45.9  \\
DDBA+DQE & 69.2 & 52.6 & 85.4 & 66.6 & 50.2 & 29.2 & 70.1 & 42.4  \\
ADBAwD+AQEwD & 74.1 & 56.2 & 84.5 & 61.5 & 54.5 & 31.1 & 68.6 & 33.7  \\
LAttDBA+LAttQE & 74.0 & 60.0 &  87.8 & 70.5 & 54.1 & 36.3 & 74.1 &  48.3 \\
\midrule
kNN \cite{cvpr2012_knn_reranking} & 71.3 & 54.7 & 83.8 & 63.2 & 49.1 & 29.2 & 66.4 & 36.7 \\
DFS \cite{cvpr2017_dfs} & 72.9 & 59.4 & 89.7 & 74.0 & 50.1 & 34.9 & 80.4 & 56.9  \\
FSR \cite{cvpr2018_fsr} & 72.7 & 59.6 & 89.6 & 73.9 & 49.6 & 34.8 & 80.2 & 56.7  \\
RDP \cite{tpami2019_rdp} & 75.2 & 55.0 & 89.7 & 70.0 & 58.8 & 33.9 & 77.9 & 48.0  \\
EIR \cite{aaai2019_eir} & 74.9 & \textbf{61.6} & 89.7 & 73.7 & 52.1 & 36.9 & 79.8 & 56.1  \\
\midrule
GSS \cite{nips2019_gss} & 78.0 & 61.5 & 88.9 & 71.8 & 60.9 & 38.4 & 76.5 & 50.1  \\
EGT \cite{cvpr2019_egt} & 74.7 & 60.1 & 87.9 & 72.6 & 51.1 & 36.2 & 76.6 & 51.3  \\
SSR \cite{nips2021_ssr} & 74.2 & 54.6 & 82.5 & 60.0 & 53.2 & 29.3 & 65.6 & 35.0  \\
CSA \cite{nips2021_csa} & 78.2 & 61.5 & 88.2 & 71.6 & 59.1 & 38.2 & 75.3 & 51.0  \\
SG \cite{iccv2023_super_global} & 71.4 & 53.9 & 83.6 & 61.5 & 49.5 & 28.8 & 67.6 & 35.8 \\
\midrule
\textbf{CAS}   & \textbf{80.7} & \textbf{61.6} & \textbf{91.0} & \textbf{75.5} & \textbf{64.8} & \textbf{39.1} & \textbf{80.7} & \textbf{59.7} \\
\bottomrule[1pt]
\end{tabular}}
\vskip -0.1in
\vspace{-0.5em}
\end{table*}

The optimal solution can be obtained by solving the KKT conditions as proved in \cref{appendix: feature enhancement}.
Consequently, the neighbor-guided smoothed similarity $\boldsymbol{\hat{F}}$ can then be obtained by applying the following smooth strategy:
\begin{equation}
    \label{eq: method feature enhancement}
    \boldsymbol{\hat{F}}_{ij} = \frac{r\boldsymbol{T}_{ij}+2\beta}{r^2+2\beta}\boldsymbol{F}_{ij} + \frac{r^2\Vert\boldsymbol{F}_i\Vert_1-r\boldsymbol{T}_i^\top\boldsymbol{F}_i}{\vert\mathcal{C}[i]\vert(r^2+2\beta)}, \quad j\in\mathcal{C}[i]
\end{equation}
Apart from increasing consistency, local neighbors can provide guidance for similarity refinement in other ways.
Primarily, the role of local neighbors can be emphasized in the diffusion process by assigning higher weights to the edges correlated with $\xi$ when constructing the nearest neighbor graph. 
Formally, the entry $\mathbbm{1}_{ij}$ of the indicator turns into $\kappa$ if $j\in\xi[i]$, where $\kappa$ is the importance weight.

Since the instances among the same local neighbors have a high probability of belonging to the same class, we aim to aggregate the similarity within the neighborhood $\xi$ to get a neighbor-enhanced representation $\boldsymbol{\widetilde{F}}$. 
This is achieved through a weighted average of the similarity matrix, where neighbors from $\mathcal{N}(i,k_2)$ are also taken into account,
\begin{equation}
    \boldsymbol{\widetilde{F}}_i = \big(\kappa\sum_{j\in \xi[i]}\boldsymbol{\hat{F}}_j/|\xi| + \sum_{j\in \mathcal{N}(i,k_2)}\boldsymbol{\hat{F}}_j/k_2\big) / (\kappa+1).
\end{equation}
The obtained similarity $\boldsymbol{\widetilde{F}}$ can be considered as the optimal representation of the data manifold within the local cluster.
To align with the global ranking task, we conduct a comprehensive propagation of similarity information to obtain $\boldsymbol{F}'$ with the transition matrix calculated by $\boldsymbol{P}=\boldsymbol{\widetilde{F}}^\top\boldsymbol{\widetilde{F}}$. 
Sparse operations can be employed to handle the transition matrix, thereby further diminishing the impact of outliers, the final similarity matrix is given by:
\begin{equation}
    \boldsymbol{F}'_i = \sum_{j}\boldsymbol{P}_{ij}\boldsymbol{\widetilde{F}}_j.
\end{equation}

\subsection{Inference}
After that, the optimized similarity matrix $\boldsymbol{F}'$ denotes the similarity among samples that can be used for retrieval. 
Formally, we jointly consider the Euclidean distance $d(i,j)$ and modified distance $d'(i,j)$ inferred by similarity matrix $\boldsymbol{F}'$ to maintain the crucial neighborhood information in Euclidean space, the final distance is represented as: 
\begin{equation}
\label{eq: distance}
    d^*(i,j)=(1-\omega) d'(i,j) + \omega d(i,j),
\end{equation}
where $\omega$ is the balance weight, and the Euclidean distance term can be replaced with diffusion-based distance to further enhance the robustness.
Since the similarity matrix $\boldsymbol{F}'$ represents the transition probability within the diffusion process, intuitively, we apply the Jensen-Shannon divergence\footnote{The variants of the formula is discussed in \cref{section: appendix sparse jensen shannon divergence}} to compute the modified distance $d'(i,j)$,
\begin{scriptsize}
\begin{equation}
\label{eq: sd}
d'(i,j) = \frac{1}{2}\bigg(\sum_{k=1}^{n}\boldsymbol{F}'_{ik}\log\Big(\frac{2\boldsymbol{F}'_{ik}}{\boldsymbol{F}'_{ik}+\boldsymbol{F}'_{jk}}\Big)+\boldsymbol{F}'_{jk}\log\Big(\frac{2\boldsymbol{F}'_{jk}}{\boldsymbol{F}'_{ik}+\boldsymbol{F}'_{jk}}\Big)\bigg).   
\end{equation}
\end{scriptsize}

\begin{table*}
    \caption{Evaluate the performance on ReID tasks, the backbone is trained on a sub-dataset (150 identities) of Market1501.}
    \label{tab: 150id reid}
    \vskip 0.1in
    \centering
    \resizebox{0.96\textwidth}{!}{
    \begin{tabular}{cccccccccccccccccccc}
    \toprule 
    \multicolumn{1}{c}{\multirow{2.4}*{Method}} & \multicolumn{2}{c}{BoT} & \multicolumn{2}{c}{OSNet} & \multicolumn{2}{c}{AGW} & \multicolumn{2}{c}{MGN} & \multicolumn{2}{c}{AdaSP} & \multicolumn{2}{c}{ABD-Net} & \multicolumn{2}{c}{TransReID} & \multicolumn{2}{c}{CLIP-ReID}  \\
    \cmidrule(lr){2-3}
    \cmidrule(lr){4-5}
    \cmidrule(lr){6-7}
    \cmidrule(lr){8-9}
    \cmidrule(lr){10-11}
    \cmidrule(lr){12-13}
    \cmidrule(lr){14-15}
    \cmidrule(lr){16-17}
    ~ & mAP & mINP & mAP & mINP & mAP & mINP & mAP  & mINP & mAP & mINP & mAP & mINP & mAP  & mINP & mAP & mINP  \\
    \midrule  
    Baseline & 54.1 & 18.1 & 57.5 & 19.3 & 57.9 & 21.5 & 65.9 & 26.0 & 60.3 & 23.2 & 64.0 & 26.7 & 72.0 & 39.3 & 63.2 & 25.9 \\
    kNN      & 60.9 & 25.4 & 63.8 & 26.9 & 64.5 & 29.5 & 72.9 & 35.4 & 66.3 & 32.8 & 70.6 & 37.3 & 77.1 & 48.7 & 70.2 & 37.3 \\
    AQE      & 69.9 & 44.1 & 71.0 & 44.1 & 71.9 & 47.5 & 81.4 & 60.9 & 73.4 & 48.6 & 77.0 & 56.9 & 81.3 & 64.2 & 77.8 & 58.1 \\
    $\alpha$QE & 69.2 & 41.2 & 69.9 & 42.7 & 71.4 & 44.8 & 80.4 & 55.2 & 73.7 & 47.9 & 74.5 & 51.9 & 79.8 & 61.1 & 77.7 & 57.9 \\
    AQEwD    & 69.5 & 43.0 & 70.9 & 42.9 & 71.8 & 46.5 & 80.9 & 59.0 & 73.5 & 47.6 & 77.5 & 54.9 & 81.5 & 63.0 & 77.3 & 56.0 \\
    RDP      & 72.1 & 50.4 & 74.3 & 50.0 & 75.6 & 55.8 & 84.0 & 67.7 & 76.7 & 54.8 & 80.2 & 62.5 & 84.5 & 70.5 & 80.1 & 62.8 \\
    SCA      & 73.6 & 54.0 & 74.4 & 51.5 & 76.3 & 57.6 & 84.5 & 69.8 & 77.1 & 56.6 & 80.9 & 64.6 & 84.1 & 71.1 & 81.0 & 65.2 \\
    $k$-recip   & 74.9 & 55.2 & 76.0 & 53.8 & 77.4 & 58.7 & 85.3 & 69.0 & 78.3 & 58.3 & 81.8 & 64.7 & 85.4 & 71.6 & 81.9 & 65.1 \\
    ECN      & 74.6 & 54.2 & 75.5 & 52.0 & 77.0 & 58.1 & 85.1 & 70.1 & 78.3 & 57.3 & 81.8 & 65.4 & 85.0 & 71.7 & 81.6 & 65.6 \\
    \midrule
    \textbf{CAS}     & \textbf{78.1} & \textbf{61.3} & \textbf{79.6} & \textbf{60.6} & \textbf{80.1} & \textbf{64.7} & \textbf{87.5} & \textbf{76.1} & \textbf{81.3} & \textbf{64.0} & \textbf{84.2} & \textbf{70.7} & \textbf{87.3} & \textbf{76.9} & \textbf{84.4} & \textbf{71.0} \\
    \bottomrule
    \end{tabular}}
    \vskip -0.1in
    \vspace{-1.0em}
\end{table*}

\section{Experiment}
\subsection{Experiment Setup}
\textbf{Datasets}: We conduct experiments on instance retrieval and object re-identification (ReID) tasks to verify the effectiveness of our proposed method.
For instance retrieval, the widely known Oxford5k \cite{cvpr2007_oxford_building} and Paris6k \cite{cvpr2008_paris} datasets are being revisited by \cite{cvpr2018_revisited}, referred to as Revisited Oxford5k (\emph{R}Oxf) and Revisited Paris6k (\emph{R}Par), respectively.
Moreover, a collection of 1 million distractor images are added to form large scale \emph{R}Oxf+1M and \emph{R}Par+1M datasets.
We also evaluate our method on object ReID datasets, including Market1501 \cite{iccv2015_market1501} and CUHK03 \cite{cvpr2014_cuhk03}.

\textbf{Evaluation Metrics}: 
The mean Average Precision (mAP) metric is used for measuring the performance of instance retrieval, and the assessment includes reporting results for Easy (E), Medium (M), and Hard (H) protocols on the \emph{R}Oxf and \emph{R}Par datasets, respectively.
For object ReID, we also consider Rank1 (CMC@1) and mean Inverse Negative Penalty (mINP) \cite{tpami2022_agw}.

\textbf{Feature Extraction}: 
For instance retrieval, maximum activation of convolutions (MAC), regional maximum activation of convolutions (R-MAC) \cite{iclr2016_rmac}, and Generalized-Mean (R-GeM) \cite{tpami2019_rtc} are used to extract image descriptors.
For object ReID, some representative methods, such as BoT \cite{cvprw2019_bot}, MGN \cite{mm2018_mgn}, AGW \cite{tpami2022_agw}, AdaSP \cite{cvpr2023_adasp_reid}, SpCL \cite{nips2020_spcl}, ISE \cite{cvpr2022_ise_reid}, OSNet \cite{tpami2022_osnet}, ABDNet \cite{iccv2019_abd_net}, TransReID \cite{iccv2021_trans_reid}, CLIP-ReID \cite{aaai2023_clip_reid} are used to extract the person descriptions.
By using a different number of identities to train the ReID model, we can construct various levels of retrieval baselines, enabling evaluation of performance in more challenging scenarios.

\subsection{Comparison with existing methods}

\textit{Comparison of Instance Retrieval.}
We first compare the proposed method with existing instance retrieval methods on \emph{R}Oxf and \emph{R}Par, and summarize the results in \cref{tab: rgem}, \cref{tab: mac} and \cref{tab: rmac}.
The compared methods consist of query expansion methods with and without database-side augmentation (DBA), including AQE, $\alpha$QE, DQE, AQEwD, SG and LAttQE; diffusion-based methods including DFS, FSR, RDP, and EIR; and the graph or attention learning-based methods including GSS, EGT, SSR and CSA. 
From those tables, we can observe that our proposed method exhibits superior performance against others with global descriptors extracted by MAC, R-MAC and R-GeM.
Among all existing methods, the diffusion-based methods are the most related to ours, and the proposed CAS obtains an obvious improvement over those diffusion-base methods, \emph{e.g.,} CAS achieves the mAP of 62.5\% and 34.1\% test with R-MAC descriptors on the medium protocol of \emph{R}Oxf, compare to the top-performing diffusion model, our approach delivers an improvement of 7.8\% and 11.9\% in mAP.
As for the \emph{R}Oxf+1M and \emph{R}Par+1M, we first perform the Euclidean search over the whole dataset, then fine arranged the top 5,000 images within each retrieval.
The superior performance demonstrates the effectiveness of our proposed CAS and its potential in re-ranking large datasets.

\begin{table}[t]
    \caption{Image retrieval performance based on MAC descriptor summarized by \cite{iclr2016_rmac}.}
    \label{tab: mac}
    \vskip 0.1in
    \centering
    \small
    \resizebox{0.4\textwidth}{!}{
    \begin{tabular}{ccccccc}
    \toprule[1pt]
     \multicolumn{1}{c}{\multirow{2.4}*{Method}} & \multicolumn{2}{c}{Easy} & \multicolumn{2}{c}{Medium} & \multicolumn{2}{c}{Hard} \\
     \cmidrule(lr){2-3}
     \cmidrule(lr){4-5}
     \cmidrule(lr){6-7}
     ~ & \makebox[0.052\textwidth][c]{\emph{R}\textbf{Oxf}} & \makebox[0.052\textwidth][c]{\emph{R}\textbf{Par}} & \makebox[0.052\textwidth][c]{\emph{R}\textbf{Oxf}} & \makebox[0.052\textwidth][c]{\emph{R}\textbf{Par}} & \makebox[0.052\textwidth][c]{\emph{R}\textbf{Oxf}} & \makebox[0.052\textwidth][c]{\emph{R}\textbf{Par}} \\
     \midrule
     MAC & 47.2 & 69.7 & 34.6 & 55.7 & 14.3 & 32.6  \\
     \midrule
     AQE & 54.4 & 80.9 & 40.6 & 67.0 & 17.1 & 45.2  \\
     $\alpha$QE & 50.3 & 77.8 & 37.1 & 64.4 & 16.3 & 43.0 \\
     DQE & 50.1 & 78.1 & 37.8 & 66.5 & 16.0 & 45.7 \\
     kNN & 56.6 & 79.7 & 41.6 & 66.5 & 17.4 & 44.5 \\
     AQEwD & 52.8 & 79.6 & 39.7 & 65.0 & 17.3 & 42.9 \\
     \midrule
     DFS & 54.6 & 83.8 & 40.6 & 74.0 & 18.8 & 58.1 \\
     FSR & 54.4 & 83.9 & 40.4 & 73.5 & 18.4 & 57.5 \\
     EIR & 57.9 & 86.9 & 44.2 & 76.8 & 22.2 & 60.5 \\
     RDP & 59.0 & 85.2 & 45.3 & 76.3 & 21.4 & 58.9 \\
     GSS & 60.0 & 87.5 & 45.4 & 76.7 & 22.8 & 59.7 \\
     \midrule
     \textbf{CAS} & \textbf{68.6} & \textbf{90.1} & \textbf{52.9} & \textbf{82.3}  & \textbf{30.4} & \textbf{68.1}  \\
     \bottomrule[1pt]
    \end{tabular}}
\vspace{-1.5em}
\end{table}

\begin{table}[t]
    \caption{Image retrieval performance based on R-MAC descriptor proposed in \cite{iclr2016_rmac}.}
    \label{tab: rmac}
    \vskip 0.1in
    \centering
    \small
    \resizebox{0.4\textwidth}{!}{
    \begin{tabular}{ccccccc}
    \toprule[1pt]
     \multicolumn{1}{c}{\multirow{2.4}*{Method}} & \multicolumn{2}{c}{Easy} & \multicolumn{2}{c}{Medium} & \multicolumn{2}{c}{Hard} \\
     \cmidrule(lr){2-3}
     \cmidrule(lr){4-5}
     \cmidrule(lr){6-7}
     ~ & \makebox[0.052\textwidth][c]{\emph{R}\textbf{Oxf}} & \makebox[0.052\textwidth][c]{\emph{R}\textbf{Par}} & \makebox[0.052\textwidth][c]{\emph{R}\textbf{Oxf}} & \makebox[0.052\textwidth][c]{\emph{R}\textbf{Par}} & \makebox[0.052\textwidth][c]{\emph{R}\textbf{Oxf}} & \makebox[0.052\textwidth][c]{\emph{R}\textbf{Par}} \\
     \midrule
     R-MAC & 61.2 & 79.3 & 40.2 & 63.8 & 10.1 & 38.2 \\
     \midrule
     AQE & 69.4 & 85.7 & 47.8 & 71.1 & 15.9 & 47.9 \\
     $\alpha$QE & 64.9 & 84.7 & 42.8 & 70.8 & 11.4 & 47.8 \\
     DQE & 65.5 & 84.9 & 45.3 & 71.9 & 15.5 & 49.1 \\
     kNN & 70.6 & 84.6 & 48.9 & 70.2 & 16.0 & 46.1 \\
     AQEwD & 70.5 & 85.9 & 48.7 & 70.7 & 15.3 & 46.9 \\
     \midrule
     DFS & 70.0 & 87.5 & 51.8 & 78.8 & 20.3 & 63.5 \\
     FSR & 69.7 & 87.3 & 51.4 & 78.1 & 20.1 & 62.6 \\
     EIR & 68.0 & 89.4 & 50.8 & 78.7 & 21.7 & 63.3 \\
     RDP & 73.7 & 88.8 & 54.3 & 79.6 & 22.2 & 61.3 \\
     GSS & 75.0 & 89.9 & 54.7 & 78.5 & 24.4 & 60.5 \\
     \midrule
     \textbf{CAS} & \textbf{82.6} & \textbf{90.0} & \textbf{62.5} & \textbf{82.5} & \textbf{34.1} & \textbf{67.4} \\
     \bottomrule[1pt]
    \end{tabular}}
\vspace{-1.5em}
\end{table}

\textit{Comparison of Object ReID.}
Object ReID is a special instance retrieval task that focuses on matching the person images captured by different camera views.
Compared to the general instance retrieval, the person has a small inter-class and large intra-class variance, leading object ReID is more challenging than the general instance retrieval.
We thus conduct the proposed CAS as a re-ranking method for instance ReID to verify its effectiveness and generalization.
As shown in \cref{tab: 150id reid}, the proposed CAS obtains a higher performance than existing re-ranking methods on all eight ReID backbones.
More comparisons of different models trained with Market1501 and CUHK03 datasets are shown in \cref{section: appendix extended results}.

\begin{table}[t]
    \caption{Effectiveness of Bidirectional Similarity Diffusion.}
    \label{tab: ablation BSD} 
    \vskip 0.05in
    \centering
    \resizebox{0.42\textwidth}{!}{\begin{tabular}{lcccccc}
    \toprule[1pt]
     \multicolumn{1}{c}{\multirow{2.4}*{Method}} & \multicolumn{3}{c}{\emph{R}Oxf(M)} & \multicolumn{3}{c}{\emph{R}Oxf(H)}\\
     \cmidrule(lr){2-4}
     \cmidrule(lr){5-7}
     ~ & MAC & R-MAC & R-GeM & MAC & R-MAC & R-GeM \\
     \midrule
        Baseline & 34.6 & 40.2 & 67.3 & 14.3 & 10.5 & 44.2 \\
        \midrule
        OSM+$k$-nn & 51.8 & 60.4 & 78.4 & 28.3 & 29.8 & 60.6 \\
        SSM+$k$-nn & 51.8 & 61.6 & 79.1 & 28.6 & 31.0 & 62.2 \\
        BSD+$k$-nn & \textbf{52.6} & \textbf{62.4} & \textbf{79.3} & \textbf{29.8} & \textbf{33.7} & \textbf{62.4} \\
        \midrule
        OSM+$k$-recip & 52.2 & 61.1 & 80.1 & 28.4 & 30.1 & 63.7 \\
        SSM+$k$-recip & 52.2 & 61.6 & 80.2 & 28.8 & 31.3 & 63.9 \\
        BSD+$k$-recip & \textbf{52.9} & \textbf{62.5} & \textbf{80.7} & \textbf{30.4} & \textbf{34.1} & \textbf{64.8} \\
    \bottomrule
    \end{tabular}}
    \vspace{-0.5em}
\end{table}

\begin{figure}[t]
    \centering
    \includegraphics[width=0.9\linewidth]{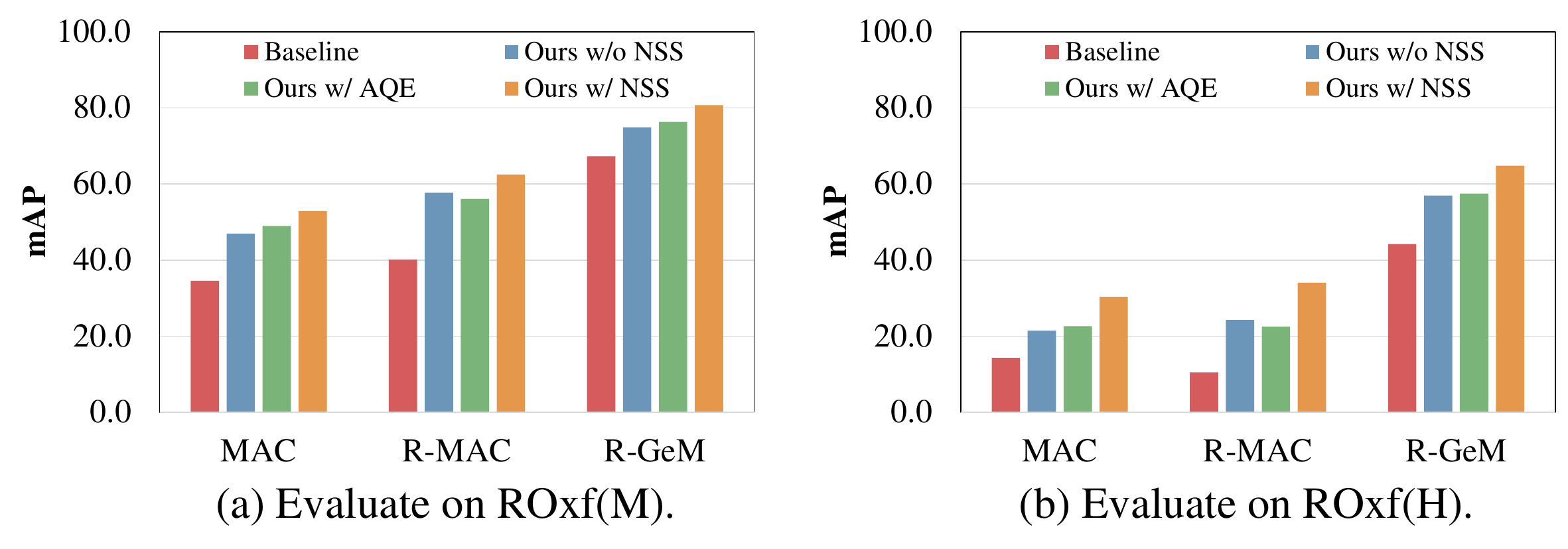}
    \caption{Effectiveness of Neighbor-guided Similarity Smooth.}
    \label{fig: ablation NSS}
\vspace{-1.5em}
\end{figure}

\subsection{Ablation Study}
\textit{Effectiveness of Bidirectional Similarity Diffusion.} 
As listed in \cref{tab: ablation BSD}, we perform several ablation studies to prove the effectiveness of our proposed Bidirectional Similarity Diffusion (BSD) strategy.
BSD comprises two crucial components, the bidirectional smooth strategy and the approximation of the local cluster.
To demonstrate the effectiveness of the bidirectional smooth strategy, we compare the proposed BSD with two existing types of similarity matrices: the original similarity matrix (OSM) in Euclidean space, and the similarity smoothed matrix (SSM) obtained by \cite{tpami2019_rdp}.
As shown in \cref{tab: ablation BSD}, the proposed BSD obtains a higher performance than OSM and SSM on both cluster generation methods such as $k$-nn and $k$-recip over three types of extractors (MAC/R-MAC/R-GeM).
For instance, in the case of $k$-reciprocal neighbors, BSD achieves an average performance of 62.5\%/34.1\%, showcasing improvements of 1.4\%/2.0\% and 0.9\%/2.8\% over OSM and SSM, respectively.
The superior performance verifies the effectiveness and rationality of Bidirectional Similarity Diffusion.

\textit{Effectiveness of Neighbor-guided Similarity Smooth.}
Moreover, we conduct the comparison to evaluate the effectiveness of our proposed Neighbor-guided Similarity Smooth strategy.
As depicted in \cref{fig: ablation NSS}, when not taking NSS into account (`\textit{w/o} NSS'), our method obtains a noticeable improvement over the baseline, which further substantiates the superiority of the Bidirectional Similarity Diffusion.
However, when compared with our full method accounting for Neighbor-guided Similarity Smooth (`\textit{w/} NSS'), there is a significant performance gap, \emph{e.g.,} the average performance gap is 5.5\% in $R$Oxf(M) and 8.8\% in $R$Oxf(H).
In addition, we draw on the idea from query expansion to fuse the similarity matrix from $k$-nearest neighbors, we then replace it with our NSS (`\textit{w/} AQE') to evaluate its performance.
The results from \cref{fig: ablation NSS} indicate that this substitution only yields limited improvement to the baseline compared with our proposed method. 
In this way, we validate the effectiveness of our proposed Neighbor-guided Similarity Smooth.

\begin{table}
    \caption{Effectiveness of the proposed distance measure.}
    \label{tab: ablation distance measure}
    \vskip 0.05in
    \centering
    \resizebox{0.42\textwidth}{!}{\begin{tabular}{lcccccc}
    \toprule[1pt]
     \multicolumn{1}{c}{\multirow{2.4}*{Method}} & \multicolumn{3}{c}{\emph{R}Oxf(M)} & \multicolumn{3}{c}{\emph{R}Oxf(H)}\\
     \cmidrule(lr){2-4}
     \cmidrule(lr){5-7}
     ~ & MAC & R-MAC & R-GeM & MAC & R-MAC & R-GeM \\
     \midrule
        Baseline & 50.6 & 59.7 & 73.2 & 27.8 & 31.1 & 54.4 \\
        Euclidean & 44.6 & 52.0 & 71.7 & 24.0 & 21.9 & 51.4 \\
        Cosine & 52.7 & 62.2 & 78.3 & 29.1 & 31.3 & 61.7 \\
        Jaccard & 52.1 & 61.6 & 79.3 & 29.5 & 31.4 & 62.3 \\
        \midrule
        JS Divergence & \textbf{52.9} & \textbf{62.5} & \textbf{80.7} & \textbf{30.4} & \textbf{34.1} & \textbf{64.8} \\
    \bottomrule
    \end{tabular}}
    \vspace{-0.5em}
\end{table}

\textit{Effectiveness of Shannon divergence.}
In Eq.\eqref{eq: sd}, the Shannon divergence is used to compute the pairwise distance after obtaining the optimized similarity matrix $\boldsymbol{F}'$. 
To verify the rationality of Shannon divergence, we also conduct the comparison by replacing Eq.\eqref{eq: sd} with Cosine, Euclidean, and Jaccard distance, as well as directly using the similarity matrix $\boldsymbol{F}'$ for retrieval (Baseline).

\begin{figure}
    \centering\setlength{\abovecaptionskip}{0.cm}
    \includegraphics[width=0.8\linewidth]{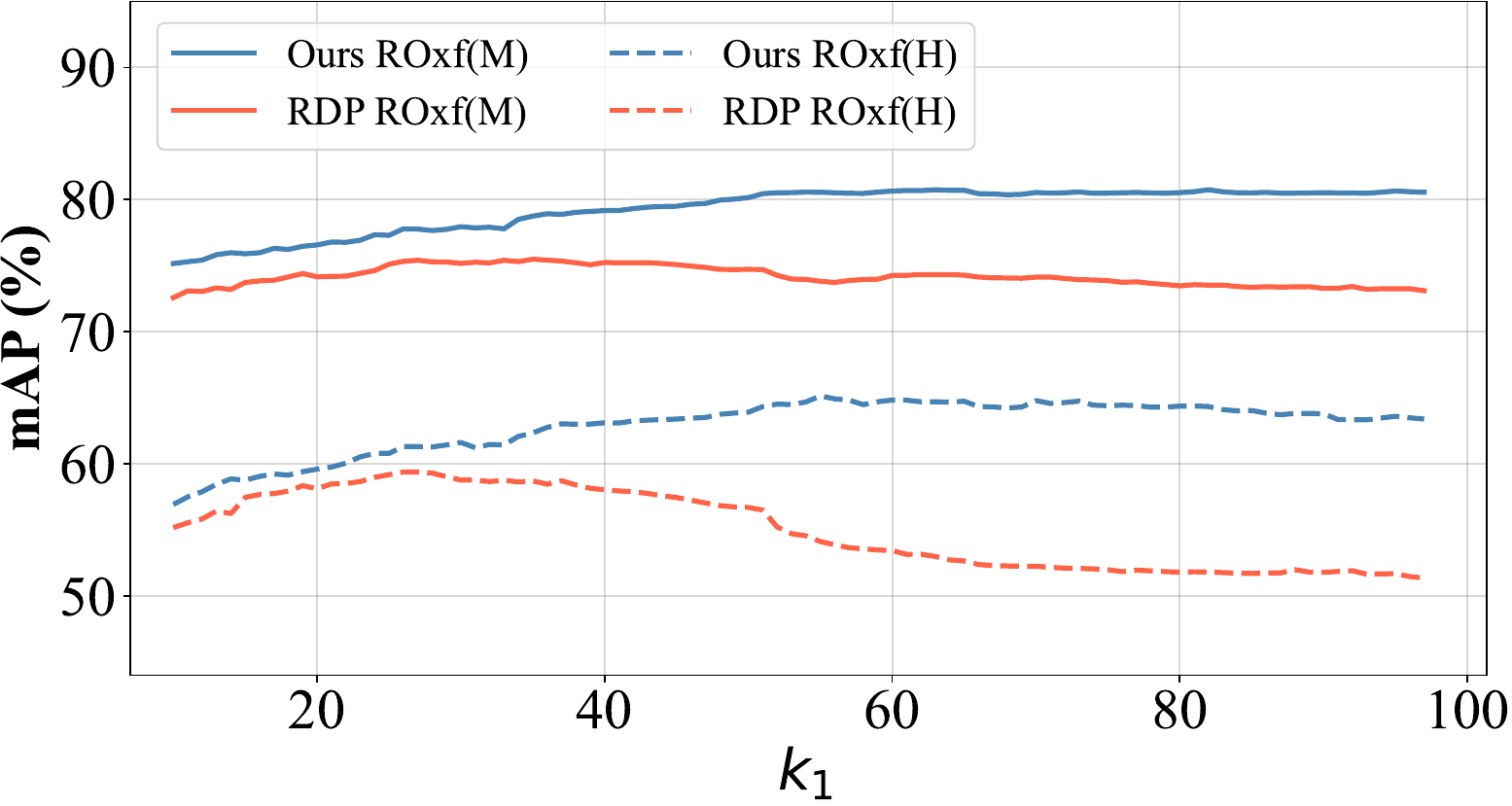}
    \vspace{-1em}
    \caption{Effect of parameter $k_1$, here we plot the sensitivity of a diffusion-based method to its hyper-parameter on the same graph.}
    \label{fig: ablation k1}
    \vspace{-1.5em}
\end{figure}

\textit{Time complexity.}
Denote the dimension of image descriptors as $d$, and the total number of re-ranking candidates as $n$.
The overall time complexity of our proposed CAS is $\mathcal{O}(n^3)$,  which is comparable to other diffusion-based methods.
Moreover, since our BSD module is highly optimized and can be parallel executed on GPU, although the time complexity is on the same level, our method still outperforms others. 
It is also noteworthy that, unlike learning-based methods such as GSS and SSR, our approach does not require training a graph neural network.
Consequently, compared to their training time of several minutes, our method only requires 1,278ms to re-rank the \emph{R}Oxford dataset.
The re-ranking latency is tested with one single RTX 3090.

\textit{Effect of $k_1$ and $k_2$}.
The hyper-parameters $k_1$ and $k_2$ are used to approximate the local cluster $\mathcal{C}$ and neighbor set $\xi$, respectively.
As shown in \cref{fig: ablation k1}, setting the hyper-parameter of the traditional diffusion-based method as $k_1$, our approach exhibits its robustness in comparison.
The further analysis of $k_2$ is shown in \cref{tab: ablation k2}.
We can observe that a balanced selection of $k_2$ yields optimal results. This is because neighbor set $\xi$ obtained through an appropriate $k_2$ can include a higher quantity and proportion of correct samples, which is crucial for adjustment.

\textit{Effect of $\omega$}.
In Eq.~\eqref{eq: distance}, the hyper-parameter $\omega$ is treated as a balance weight to fuse the original Euclidean distance and the Jensen-Shannon divergence.
As shown in Table~\ref{tab: ablation omega}, the optimal result is attained when $\omega$ is set to 0.2 and 0.3.
This achievement is attributed to our incorporation of the original distance, thereby improving the robustness of the retrieval process.

\begin{table}[t]
\vspace{-0.5em}
    \caption{Analysis of time complexity.}
    \label{tab: time}
    \centering
    \resizebox{0.4\textwidth}{!}{\begin{tabular}{ccc}
        \toprule
        Method & Time Complexity & Re-ranking Latency (ms) \\
        \midrule
        $\alpha$QE & $\mathcal{O}(n^2d)$ & 121 \\
        $k$-recip & $\mathcal{O}(n^3)$ & 8,524 \\
        DFS & $\mathcal{O}(n^3)$ & 1,857 \\
        RDP & $\mathcal{O}(n^3)$ & 6,018 \\
        GSS & - & $>$ 5 min \\
        \midrule
        CAS & $\mathcal{O}(n^3)$ & 1,278 \\
        \bottomrule
    \end{tabular}}
\vspace{-1.5em}
\end{table}

\begin{table}[t]
    \caption{Effect of parameter $k_2$.}
    \label{tab: ablation k2}
    \centering
    \resizebox{0.44\textwidth}{!}{\begin{tabular}{ccccccccccc}
        \toprule
        $k_2$ & 1 & 2 & 3  & 4  & 5  & 6  & 7 & 8  & 9  & 10  \\
        \midrule
        \emph{R}Oxf(M)  & 78.14 & 78.81 & 79.7 & \textbf{80.8} & \textbf{80.8} & 80.7 & 80.5 & 79.2 & 78.3 & 78.0 \\
        \emph{R}Oxf(H)  & 61.27 & 62.44 & 64.0 & \textbf{65.6} & 65.3 & 64.8 & 64.9 & 62.7 & 61.2 & 61.2 \\
        \bottomrule
    \end{tabular}}
\vspace{-1.5em}
\end{table}

\begin{table}[t]
\vspace{-0.5em}
    \caption{Effect of parameter $\omega$.}
    \label{tab: ablation omega}
    \centering
    \resizebox{0.44\textwidth}{!}{\begin{tabular}{ccccccccccc}
        \toprule
        $\omega$ & 0.01 & 0.05 & 0.1  & 0.2  & 0.3  & 0.4  & 0.5  & 0.6  & 0.7  & 0.8  \\
        \midrule
        \emph{R}Oxf(M)  & 79.1 & 80.0 & 80.5 & \textbf{80.7} & \textbf{80.7} & 80.5 & 80.1 & 79.4 & 78.4 & 76.5 \\
        \emph{R}Oxf(H)  & 64.2 & 64.6 & 65.2 & \textbf{65.3} & 64.8 & 63.9 & 63.0 & 61.8 & 60.4 & 58.0 \\
        \bottomrule
    \end{tabular}}
\vspace{-1em}
\end{table}

\section{Conclusion}

The existing diffusion-based method suffers from the influence affected by nearby manifolds, which restricts its performance.
To address this issue, we propose a novel Cluster-Aware Similarity (CAS) diffusion method for instance retrieval.
CAS confines diffusion within a local cluster and utilizes a neighbor set to refine the obtained similarity matrix.
Extensive evaluation of several benchmarks demonstrates the effectiveness of the proposed CAS in boosting the retrieval performance, indicating that CAS can effectively suppress the negative impacts from neighboring manifolds in the diffusion process.

Nevertheless, there is still room for improvement in providing a more accurate estimation of the local cluster.
In the future, we aim to explore a more effective method for local cluster diffusion and enable the model to autonomously adjust its hyper-parameters.

\section*{Acknowledgements}
This work was supported by National Science and Technology Major Project (2021ZD0112202), National Natural Science Foundation of China (62376268, U23A20387, 62036012, U21B2044, 62376196), and Beijing Natural Science Foundation (4222039).

\section*{Impact Statement}
This paper presents work whose goal is to advance the field of 
Machine Learning. There are many potential societal consequences 
of our work, none of which we feel must be specifically highlighted here.

\bibliography{reference}
\bibliographystyle{icml2024}

\newpage
\appendix
\onecolumn
\def\eg{\emph{e.g.}} \def\Eg{\emph{E.g.}}
\def\ie{\emph{i.e.}} \def\Ie{\emph{I.e.}}

\icmltitle{Cluster-Aware Similarity Diffusion for Instance Retrieval}

\section{Bidirectional Similarity Diffusion}
\subsection{Basic Similarity Diffusion Process}
\label{section: appendix basic similarity diffusion process}

In this section, we will first introduce a Basic Similarity Diffusion Process. 
Follow the settings in \cite{nips2003_learning_with_local_and_global_consistency, cvpr2013_diffusion_processes, cvpr2017_dfs}, 
the similarities form $x_i$ and $x_j$ to other points in the data manifold are constrained by the same affinity weight $\boldsymbol{W}_{ij}$, where $\boldsymbol{W}$ is explicitly designed to be a symmetric matrix.
In addition, we expect the final obtained $\boldsymbol{F}$ to be under the supervision of a self-similarity matrix $\boldsymbol{I}$, thus we introduce a regularization term at the end. The optimal value is obtained by minimizing the following objective function $J$:
\begin{equation}
    \label{eq: appendix basic objective function}
    \begin{aligned}
    J = &\frac{1}{2}\sum_{i,j=1}^{n} \boldsymbol{W}_{ij}\Big\Vert\frac{\boldsymbol{F}_{i*}}{\sqrt{\boldsymbol{D}_{ii}}} - \frac{\boldsymbol{F}_{j*}}{\sqrt{\boldsymbol{D}_{jj}}}\Big\Vert_2^2 + \mu\Vert\boldsymbol{F}-\boldsymbol{I}\Vert^2_2\\
    =&\frac{1}{2}\sum_{i,j=1}^{n} \boldsymbol{W}_{ij} \Big(\frac{\boldsymbol{F}_{i*}(\boldsymbol{F}_{i*})^\top}{\boldsymbol{D}_{ii}} - 2\frac{\boldsymbol{F}_{i*}(\boldsymbol{F}_{j*})^\top}{\sqrt{\boldsymbol{D}_{ii}\boldsymbol{D}_{jj}}} + \frac{\boldsymbol{F}_{j*}(\boldsymbol{F}_{j*})^\top}{\boldsymbol{D}_{jj}}\Big) + \mu\Vert\boldsymbol{F}-\boldsymbol{I}\Vert^2_2 \\
    = & \frac{1}{2} \big(2tr(\boldsymbol{F}\boldsymbol{F}^\top) - 2tr(\boldsymbol{S}^\top\boldsymbol{F}\boldsymbol{F}^\top)\big) + \mu tr\big((\boldsymbol{F}-\boldsymbol{I})(\boldsymbol{F}-\boldsymbol{I})^\top\big), \\
    \end{aligned}
\end{equation}
where $\boldsymbol{F}_{i*}$ and $\boldsymbol{F}_{*j}$ are used to denote the $i$-th row and $j$-th column of $\boldsymbol{F}$ respectively, and $tr(\cdot)$ represents the trace operator that return the summation of the diagonal items in a matrix.
In addition, define the normalized matrix $\boldsymbol{S}$ as $\boldsymbol{S}=\boldsymbol{D}^{-1/2}\boldsymbol{W}\boldsymbol{D}^{-1/2}$.
To solve the minimization problem, we first show that the objective function $J$ is convex. 
By expanding the last trace item and merging it with the previous ones, we can get the following equation:
\begin{equation}
    \begin{aligned}
    J &= tr(\boldsymbol{F}\boldsymbol{F}^\top) - tr(\boldsymbol{S}^\top\boldsymbol{F}\boldsymbol{F}^\top) + \mu tr(\boldsymbol{F}\boldsymbol{F}^\top) -2\mu tr(\boldsymbol{F})+\mu tr(\boldsymbol{I}) \\
    &=tr\big(\boldsymbol{F}^\top(\mu \boldsymbol{I}+\boldsymbol{I}-\boldsymbol{S})\boldsymbol{F}\big)-2\mu tr(\boldsymbol{F})+\mu tr(\boldsymbol{I}).
    \end{aligned}
\end{equation}
Later in \cref{section: appendix bidirectional similarity diffusion process}, we will prove that for a normalized graph matrix $\boldsymbol{S}$, its eigenvalues are no larger than 1.
We can conclude that $(\mu+1)\boldsymbol{I}-\boldsymbol{S}$ is positive definite, thus there exists a matrix $\boldsymbol{M}$ such that $(\mu+1)\boldsymbol{I}-\boldsymbol{S}=\boldsymbol{M}\boldsymbol{M}^\top$. The objective function $J$ turns out to be:
\begin{equation}
    \begin{aligned}                                 
    J&=tr(\boldsymbol{F}^\top\boldsymbol{M}\boldsymbol{M}^\top\boldsymbol{F}) - 2\mu tr(\boldsymbol{F}) + \mu tr(\boldsymbol{I}) \\
    &= \Vert\boldsymbol{M}^\top\boldsymbol{F}\Vert_F  - 2\mu tr(\boldsymbol{F}) + \mu tr(\boldsymbol{I}).
    \end{aligned}
\end{equation}
The first item denotes the Frobenius norm of $\boldsymbol{M}^\top\boldsymbol{F}$, which is strictly convex, the remaining parts can be viewed as affine functions, thus $J$ is convex. Take the derivative of $J$ to variable $\boldsymbol{F}$, we can get:
\begin{equation}
    \begin{aligned}
        \nabla_{\boldsymbol{F}} J &= 2(\mu+1)\boldsymbol{F}-(\boldsymbol{S}+\boldsymbol{S}^\top)\boldsymbol{F} - 2\mu \boldsymbol{I}. \\
    \end{aligned}
\end{equation}
In the basic settings, both $\boldsymbol{W}$ and $\boldsymbol{S}$ are symmetric, set the above derivation to 0, we can conclude that the optimum value of $\boldsymbol{F}$ is the solution to the following equation:
\begin{equation}
    \begin{aligned}
        \big((\mu+1)\boldsymbol{I}-\boldsymbol{S}\big)\boldsymbol{F} = \mu \boldsymbol{I},
    \end{aligned}
\end{equation}
by substituting $\alpha=\frac{1}{\mu+1}$, we can simplify the optimal result as below:
\begin{equation}
    \begin{aligned}
        \boldsymbol{F}^* = (1-\alpha)(\boldsymbol{I}-\alpha \boldsymbol{S})^{-1}\boldsymbol{I}.
    \end{aligned}
\end{equation}
The resulting matrix derived from the Basic Similarity Process captures more manifold information, providing a more dependable foundation for ranking.
However, computing the inverse of matrix $\boldsymbol{I}-\alpha\boldsymbol{S}$ is a process with very high time complexity.
Zhou et al. \yrcite{nips2003_ranking_on_data_manifolds} proved that the optimal result can be iteratively approached, reducing the time complexity to $\mathcal{O}(n^3)$, where $n$ is the dimension of matrix $\boldsymbol{S}$. 
Moreover, following \cite{cvpr2017_dfs, cvpr2018_fsr}, the iterative time can be further reduced by utilizing the conjugate gradient method.

\subsection{Bidirectional Similarity Diffusion Process}
\label{section: appendix bidirectional similarity diffusion process}

To reveal the underlying manifold information in the Euclidean space, we first utilize the initial feature embedding to construct an affinity graph $\mathcal{G}=\{\mathcal{V},\mathcal{E}\}$, which explicitly represents the data manifold by connecting the nearest neighbors with a weight matrix $\boldsymbol{W}$.
The diffusion process aims to obtain a manifold-aware similarity matrix by performing the information propagation within the affinity graph.
Unlike the previous work \cite{tpami2019_rdp, tpami2013_tpg, nips2012_fusion_with_diffusion} that conducts diffusion in a hypergraph, the Bidirectional Similarity Diffusion Process only utilizes direct neighbors to regularize pairwise similarity weights, thereby better mitigating the influence of outliers.
Intuitively, we introduce a reverse smoothness term to ensure the symmetry of the resulting similarity matrix, \emph{i.e.}, the affinity weight $\boldsymbol{W}_{ij}$ is used to constrain both the forward pair of similarities $\boldsymbol{F}_{ki}$ and $\boldsymbol{F}_{kj}$, as well as the reverse pair $\boldsymbol{F}_{ik}$ and $\boldsymbol{F}_{jk}$.
In this section, we will reformulate the original optimization problem of the Bidirectional Similarity Diffusion Process into a matrix form and demonstrate its equivalence to solving a corresponding Lyapunov equation.
The objective function is defined as:
\begin{equation}
\label{eq: appendix generalized objective function}
    \begin{aligned}
    \min_{\boldsymbol{F}} \underbrace{\frac{1}{4}\sum_{k=1}^{n}\sum_{i,j=1}^{n} \bigg(\boldsymbol{W}_{ij}\Big(\frac{\boldsymbol{F}_{ki}}{\sqrt{\boldsymbol{D}_{ii}}} - \frac{\boldsymbol{F}_{kj}}{\sqrt{\boldsymbol{D}_{jj}}}\Big)^2 + \boldsymbol{W}_{ij}\Big(\frac{\boldsymbol{F}_{ik}}{\sqrt{\boldsymbol{D}_{ii}}} - \frac{\boldsymbol{F}_{jk}}{\sqrt{\boldsymbol{D}_{jj}}}\Big)^2\bigg)}_{\text{smoothness}} + \underbrace{\vphantom{\frac{1}{4}\sum_{k=1}^{n}\sum_{i,j=1}^{n} \bigg(\boldsymbol{W}_{ij}\Big(\frac{\boldsymbol{F}_{ki}}{\sqrt{\boldsymbol{D}_{ii}}} - \frac{\boldsymbol{F}_{kj}}{\sqrt{\boldsymbol{D}_{jj}}}\Big)^2 +\bigg)}{\mu\Vert\boldsymbol{F}-\boldsymbol{E}\Vert^2_F}}_{\text{regularization}},
    \end{aligned}
\end{equation}
where the left and right side in the expression are refereed to as the smoothness term and regularization term respectively.
Similar to the definition in \cref{section: appendix basic similarity diffusion process}, $\boldsymbol{D}$ is the diagonal matrix with its element equal to the summation of the corresponding row in the weight matrix $\boldsymbol{W}$, \emph{i.e.}, $\boldsymbol{D}_{ii}=\sum_j\boldsymbol{W}_{ij}$.
The regularization term is weighted by the hyper-parameter $\mu$ and the semi-positive matrix $\boldsymbol{E}$ is introduced to prevent the objective similarity matrix $\boldsymbol{F}$ from becoming excessively smooth.
In order to convert the expression into matrix format, we involve a new identity matrix $\boldsymbol{I}$ to streamline the derivation, the smoothness term can then be transformed into:
\begin{equation}
    \label{eq: appendix smooth term}
    \begin{aligned}
    \frac{1}{4}\sum_{k,l=0}^{n}\sum_{i,j=0}^{n}\bigg( \underbrace{\boldsymbol{W}_{ij}\boldsymbol{I}_{kl}\Big(\frac{\boldsymbol{F}_{ki}}{\sqrt{\boldsymbol{D}_{ii}}} - \frac{\boldsymbol{F}_{lj}}{\sqrt{\boldsymbol{D}_{jj}}}\Big)^2}_{\text{part 1}} + \underbrace{\boldsymbol{I}_{kl}\boldsymbol{W}_{ij}\Big(\frac{\boldsymbol{F}_{ik}}{\sqrt{\boldsymbol{D}_{ii}}} - \frac{\boldsymbol{F}_{jl}}{\sqrt{\boldsymbol{D}_{jj}}}\Big)^2}_{\text{part 2}}\bigg).
    \end{aligned}
\end{equation}
For further transformation, we need to introduce the vectorization operator $vec(\cdot)$ and the Kronecker product $\otimes$.
The vectorization operator $vec(\cdot)$ can transform a matrix into a column vector by stacking its columns sequentially, while the Kronecker product $\otimes$ can generate a larger block matrix by multiplying each element of the first matrix with the entire second matrix.
Moreover, define $\alpha\equiv n(i-1)+k$, $\beta\equiv n(j-1)+l$, $\mathbb{W}^{(1)}=\boldsymbol{W}\otimes \boldsymbol{I}$ and $\mathbb{D}^{(1)}=\boldsymbol{D}\otimes \boldsymbol{I}$ for the first part in Eq.~\eqref{eq: appendix smooth term}, $\gamma\equiv n(k-1)+i$ and $\delta\equiv n(l-1)+j$, $\mathbb{W}^{(2)}=\boldsymbol{I}\otimes \boldsymbol{W}$ and $\mathbb{D}^{(2)}=\boldsymbol{I}\otimes \boldsymbol{D}$ for the second part in Eq.~\eqref{eq: appendix smooth term}. 
In addition to this, define the normalized matrix as $\boldsymbol{S}=\boldsymbol{D}^{-1/2}\boldsymbol{W}\boldsymbol{D}^{-1/2}$, $\mathbb{S}^{(1)}=\boldsymbol{S}\otimes \boldsymbol{I}$ and $\mathbb{S}^{(2)}=\boldsymbol{I}\otimes \boldsymbol{S}$. 
Then we can reformulate the smoothness term into matrix form as follows:
\begin{equation}
    \label{eq: appendix matrix smooth term}
    \begin{aligned}
    &\frac{1}{4}\sum_{\alpha,\beta=1}^{n^2} \mathbb{W}^{(1)}_{\alpha\beta} \Big(\frac{vec(\boldsymbol{F})_\alpha}{\sqrt{\mathbb{D}^{(1)}_{\alpha\alpha}}} - \frac{vec(\boldsymbol{F})_\beta}{\sqrt{\mathbb{D}^{(1)}_{\beta\beta}}}\Big)^2 + 
    \frac{1}{4}\sum_{\gamma,\delta=1}^{n^2} \mathbb{W}^{(2)}_{\gamma\delta} \Big(\frac{vec(\boldsymbol{F})_\gamma}{\sqrt{\mathbb{D}^{(2)}_{\gamma\gamma}}} - \frac{vec(\boldsymbol{F})_\delta}{\sqrt{\mathbb{D}^{(2)}_{\delta\delta}}}\Big)^2\\
    =&\frac{1}{4}\sum_{\alpha,\beta=1}^{n^2} \mathbb{W}^{(1)}_{\alpha\beta}\frac{vec(\boldsymbol{F})_\alpha^2}{\mathbb{D}^{(1)}_{\alpha\alpha}}
    + \frac{1}{4}\sum_{\alpha,\beta=1}^{n^2} \mathbb{W}^{(1)}_{\alpha\beta}\frac{vec(\boldsymbol{F})_\beta^2}{\mathbb{D}^{(1)}_{\beta\beta}} - \frac12\sum_{\alpha,\beta=1}^{n^2} vec(\boldsymbol{F})_\alpha \frac{\mathbb{W}^{(1)}_{\alpha\beta}}{\sqrt{\mathbb{D}^{(1)}_{\alpha\alpha}\mathbb{D}^{(1)}_{\beta\beta}}} vec(\boldsymbol{F})_\beta \\
    &\frac{1}{4}\sum_{\gamma,\delta=1}^{n^2} \mathbb{W}^{(2)}_{\gamma\delta}\frac{vec(\boldsymbol{F})_\gamma^2}{\mathbb{D}^{(2)}_{\gamma\gamma}}
    + \frac{1}{4}\sum_{\gamma,\delta=1}^{n^2} \mathbb{W}^{(2)}_{\gamma\delta}\frac{vec(\boldsymbol{F})_\delta^2}{\mathbb{D}^{(2)}_{\delta\delta}} - \frac12\sum_{\gamma,\delta=1}^{n^2} vec(\boldsymbol{F})_\gamma \frac{\mathbb{W}^{(2)}_{\gamma\delta}}{\sqrt{\mathbb{D}^{(2)}_{\gamma\gamma}\mathbb{D}^{(2)}_{\delta\delta}}} vec(\boldsymbol{F})_\delta\\
    =&vec(\boldsymbol{F})^\top vec(\boldsymbol{F}) - \frac12vec(\boldsymbol{F})^\top\big((\mathbb{D}^{(1)})^{-1/2}\mathbb{W}^{(1)}(\mathbb{D}^{(1)})^{-1/2} + (\mathbb{D}^{(2)})^{-1/2}\mathbb{W}^{(2)}(\mathbb{D}^{(2)})^{-1/2}\big)vec(\boldsymbol{F})\\
    =&vec(\boldsymbol{F})^\top\big(\mathbb{I}-\frac12\mathbb{S}^{(1)}-\frac12\mathbb{S}^{(2)}\big)vec(\boldsymbol{F}).
    \end{aligned}
\end{equation}
The above inference utilizes the following facts:
\begin{enumerate}
    \item $vec(\boldsymbol{F})_\alpha = \boldsymbol{F}_{ki}$, $vec(\boldsymbol{F})_\beta = \boldsymbol{F}_{lj}$ in part 1, $vec(\boldsymbol{F})_\gamma = \boldsymbol{F}_{ik}$ and $vec(\boldsymbol{F})_\delta = \boldsymbol{F}_{jl}$ in part 2.
    \item $\mathbb{W}^{(1)}_{\alpha\beta}=\boldsymbol{W}_{ij}\boldsymbol{I}_{kl}$, $\mathbb{D}^{(1)}_{\alpha\alpha}=\boldsymbol{D}_{ii}$ and $\mathbb{D}^{(1)}_{\beta\beta}=\boldsymbol{D}_{jj}$ in part 1, $\mathbb{W}^{(2)}_{\gamma\delta}=\boldsymbol{I}_{kl}\boldsymbol{W}_{ij}$, $\mathbb{D}^{(2)}_{\gamma\gamma}=\boldsymbol{D}_{ii}$ and $\mathbb{D}^{(2)}_{\delta\delta}=\boldsymbol{D}_{jj}$ in part 2.
    \item $\sum_{\beta=1}^{n^2}\mathbb{W}^{(1)}_{\alpha\beta}=\mathbb{D}^{(1)}_{\alpha\alpha}$, $\sum_{\alpha=1}^{n^2}\mathbb{W}^{(1)}_{\alpha\beta}=\mathbb{D}^{(1)}_{\beta\beta}$, $\sum_{\delta=1}^{n^2}\mathbb{W}^{(2)}_{\gamma\delta}=\mathbb{D}^{(2)}_{\gamma\gamma}$ and $\sum_{\gamma=1}^{n^2}\mathbb{W}^{(2)}_{\gamma\delta}=\mathbb{D}^{(2)}_{\delta\delta}$ since:
    \begin{equation*}
        \begin{aligned}
        \sum_{\beta=1}^{n^2} \mathbb{W}^{(1)}_{\alpha\beta} = \sum_{j=1}^{n}\boldsymbol{W}_{ij}\sum_{l=1}^{n}\boldsymbol{I}_{kl} = \boldsymbol{D}_{ii} = \mathbb{D}^{(1)}_{\alpha\alpha}, \qquad \sum_{\alpha=1}^{n^2} \mathbb{W}^{(1)}_{\alpha\beta} = \sum_{i=1}^{n}\boldsymbol{W}_{ij}\sum_{k=1}^{n}\boldsymbol{I}_{kl} = \boldsymbol{D}_{jj} = \mathbb{D}^{(1)}_{\beta\beta}, \\
        \sum_{\delta=1}^{n^2} \mathbb{W}^{(2)}_{\gamma\delta} = \sum_{l=1}^{n}\boldsymbol{I}_{kl}\sum_{j=1}^{n}\boldsymbol{W}_{ij} = \boldsymbol{D}_{ii} = \mathbb{D}^{(2)}_{\gamma\gamma}, \qquad \sum_{\gamma=1}^{n^2} \mathbb{W}^{(2)}_{\gamma\delta} = \sum_{k=1}^{n}\boldsymbol{I}_{kl}\sum_{i=1}^{n}\boldsymbol{W}_{ij} = \boldsymbol{D}_{jj} = \mathbb{D}^{(2)}_{\delta\delta}.
        \end{aligned}
    \end{equation*}
    \item $\mathbb{S}^{(1)}=(\mathbb{D}^{(1)})^{-1/2}\mathbb{W}^{(1)}(\mathbb{D}^{(1)})^{-1/2}$ and $\mathbb{S}^{(2)}=(\mathbb{D}^{(2)})^{-1/2}\mathbb{W}^{(2)}(\mathbb{D}^{(2)})^{-1/2}$ since:
     \begin{equation*}
        \begin{aligned}
        \mathbb{S}^{(1)}_{\alpha\beta} &= \boldsymbol{S}_{ij}\boldsymbol{I}_{kl} = \boldsymbol{D}_{ii}^{-1/2}\boldsymbol{W}_{ij}\boldsymbol{D}_{jj}^{-1/2}\boldsymbol{I}_{kl} &\qquad \qquad \mathbb{S}^{(2)}_{\gamma\delta} &= \boldsymbol{I}_{kl}\boldsymbol{S}_{ij} = \boldsymbol{I}_{kl}\boldsymbol{D}_{ii}^{-1/2}\boldsymbol{W}_{ij}\boldsymbol{D}_{jj}^{-1/2} \\
        &= \boldsymbol{D}_{ii}^{-1/2}\boldsymbol{W}_{ij}\boldsymbol{I}_{kl}\boldsymbol{D}_{jj}^{-1/2} & &= \boldsymbol{D}_{ii}^{-1/2}\boldsymbol{I}_{kl}\boldsymbol{W}_{ij}\boldsymbol{D}_{jj}^{-1/2} \\
        &= (\mathbb{D}^{(1)}_{\alpha\alpha})^{-1/2}(\mathbb{W}^{(1)})_{\alpha\beta}(\mathbb{D}^{(1)}_{\beta\beta})^{-1/2}, & &= (\mathbb{D}^{(2)}_{\gamma\gamma})^{-1/2}\mathbb{W}^{(2)}_{\gamma\delta}(\mathbb{D}^{(2)}_{\delta\delta})^{-1/2}.\\
        \end{aligned}
    \end{equation*}
\end{enumerate}
The Frobenius regularization term in Eq.~\eqref{eq: appendix generalized objective function} is equivalent to the squared norm of $vec(\boldsymbol{F}-\boldsymbol{E})$.
By jointly considering the smoothness term and the regularization term, the objective function can be formulated as:
\begin{equation}
    \begin{aligned}
    \min_{\boldsymbol{F}} vec(\boldsymbol{F})^T\big(\mathbb{I}-\frac12\mathbb{S}^{(1)}-\frac12\mathbb{S}^{(2)}\big)vec(\boldsymbol{F})+\mu\Vert vec(\boldsymbol{F}-\boldsymbol{E})\Vert^2_2.
    \end{aligned}
\end{equation}
\begin{lemma}
    \label{lemma: appendix radius constrain}
    Let $\boldsymbol{A}\in\mathbb{R}^{n\times n}$, the spectral radius of $\boldsymbol{A}$ is denoted as $\rho(\boldsymbol{A})=\max\{|\lambda|,\lambda\in\sigma(\boldsymbol{A})\}$, where $\sigma(\boldsymbol{A})$ is the spectrum of $\boldsymbol{A}$ that represents the set of all the eigenvalues. 
    Let $\Vert\cdot\Vert$ be a matrix norm on $\mathbb{R}^{n\times n}$, given a square matrix $\boldsymbol{A}\in\mathbb{R}^{n\times n}$, $\lambda$ is an arbitrary eigenvalue of $\boldsymbol{A}$, then we have $|\lambda|\leq\rho(\boldsymbol{A})\leq\Vert\boldsymbol{A}\Vert$.
\end{lemma}
\begin{lemma}
    \label{lemma: appendix kronecker eigenvalue}
    Let $\boldsymbol{A}\in\mathbb{R}^{m\times m}$, $\boldsymbol{B}\in\mathbb{R}^{n\times n}$, denote $\{\lambda_i,\boldsymbol{x}_i\}_{i=1}^{m}$ and $\{\mu_i,\boldsymbol{y}_i\}_{i=1}^{n}$ as the eigen-pairs of $\boldsymbol{A}$ and $\boldsymbol{B}$ respectively. The set of $mn$ eigen-pairs of $\boldsymbol{A}\otimes \boldsymbol{B}$ is given by
    \begin{equation*}
        \{\lambda_i\mu_j, \boldsymbol{x}_i\otimes \boldsymbol{y}_j\}_{i=1,\dots,m,\ j=1,\dots n}.
    \end{equation*}
\end{lemma}
Denote the objective function as $J$, then we will prove that it is convex, which allows us to easily find the optimal solution at the point where the partial derivative is zero.
Consider matrix $\boldsymbol{D}^{-1}\boldsymbol{W}$, since $\boldsymbol{D}$ is a diagonal matrix with its $i$-th element the sum of the corresponding $i$-th row of matrix $\boldsymbol{W}$, we can easily obtain that $\Vert\boldsymbol{D}^{-1}\boldsymbol{W}\Vert_{\infty}=1$. 
According to \cref{lemma: appendix radius constrain}, it is obviously that $\rho(\boldsymbol{D}^{-1}\boldsymbol{W})\leq1$. 
As for the matrix $\boldsymbol{S}=\boldsymbol{D}^{-1/2}\boldsymbol{W}\boldsymbol{D}^{-1/2}$ we are concerned about, we can rewrite it as $\boldsymbol{D}^{1/2}\boldsymbol{D}^{-1}\boldsymbol{W}\boldsymbol{D}^{-1/2}$, which means $\boldsymbol{S}\sim \boldsymbol{D}^{-1}W$. 
Since two similar matrices share the same eigenvalues, we can conclude that $\rho(\boldsymbol{S})\leq1$. 
By applying \cref{lemma: appendix kronecker eigenvalue}, the spectral radius of the Kronecker product $\mathbb{S}^{(1)}=\boldsymbol{S}\otimes \boldsymbol{I}$ and $\mathbb{S}^{(2)}=\boldsymbol{I}\otimes \boldsymbol{S}$ is no larger than $1$, \emph{i.e.}, $\rho(\mathbb{S}^{(1)})\leq1$, $\rho(\mathbb{S}^{(2)})\leq1$.

The Hessian matrix $\boldsymbol{H}$ of the objective function is $2(\mu+1)\mathbb{I}-\Bar{\mathbb{S}}^{(1)}-\Bar{\mathbb{S}}^{(2)}$, where $2\Bar{\mathbb{S}}^{(1)}=\mathbb{S}^{(1)}+(\mathbb{S}^{(1)})^\top$ and $2\Bar{\mathbb{S}}^{(2)}=\mathbb{S}^{(2)}+(\mathbb{S}^{(2)})^\top$. 
Since $\mu>0$ and $\rho(\mathbb{S})\leq1$, we can observe that all the eigenvalue of $\boldsymbol{H}$ is larger than 0, which means the Hessian matrix $\boldsymbol{H}$ is positive definite. 
The positive definite Hessian matrix implies that the objective function is convex, in this case, we can solve the optimal problem by taking the partial derivative of $vec(\boldsymbol{F})$, that is:
\begin{equation}
    \label{eq: appendix partial derivative}
    \nabla_{vec(\boldsymbol{F})}J = (2\mathbb{I}-\Bar{\mathbb{S}}^{(1)}-\Bar{\mathbb{S}}^{(2)})vec(\boldsymbol{F}) + 2\mu(vec(\boldsymbol{F}-\boldsymbol{E})),
\end{equation}
the optimal value is attained when the partial derivative is equal to zero, thus we can get the solution as:
\begin{equation}
    vec(\boldsymbol{F}^*) = \frac{2\mu}{\mu+1}\Big(2\mathbb{I} - \frac{1}{\mu+1}\Bar{\mathbb{S}}^{(1)} - \frac{1}{\mu+1}\Bar{\mathbb{S}}^{(2)}\Big)^{-1}vec(\boldsymbol{E}),
\end{equation}
by substituting $\alpha=\frac{1}{\mu+1}$ and $\Bar{\mathbb{S}}=(\Bar{\mathbb{S}}^{(1)}+\Bar{\mathbb{S}}^{(2)})/2$, we can obtain the following simplified expression as:
\begin{equation}
    \label{eq: appendix optimal solution}
    \boldsymbol{F}^* = (1-\alpha)vec^{-1}\big((\mathbb{I}-\alpha\Bar{\mathbb{S}})^{-1}vec(\boldsymbol{E})\big).
\end{equation}
\begin{lemma}
    \label{lemma: appendix vectorization kronecker product}
    Let $\boldsymbol{A}\in\mathbb{R}^{m\times n}$, $\boldsymbol{X}\in\mathbb{R}^{n\times p}$ and $\boldsymbol{B}\in\mathbb{R}^{p\times q}$ respectively, then
    \begin{equation*}
        vec(\boldsymbol{AXB}) = (\boldsymbol{B}^\top\otimes \boldsymbol{A})vec(\boldsymbol{X}).
    \end{equation*}
\end{lemma}
Additionally, by utilizing the relationship provided by \cref{lemma: appendix vectorization kronecker product} and maintaining the partial derivatives given in Eq.~\eqref{eq: appendix partial derivative} as zero, we can derive the following expression:
\begin{equation}
    \begin{aligned}
        2\boldsymbol{F} - \boldsymbol{F}{\boldsymbol{\Bar{S}}} - {\boldsymbol{\Bar{S}}}\boldsymbol{F} + 2\mu(\boldsymbol{F}-\boldsymbol{E}) = 0,
    \end{aligned}
\end{equation}
where $\Bar{\boldsymbol{S}}=(\boldsymbol{S}+\boldsymbol{S}^\top)/2$. By performing some simple substitutions, we can obtain that the optimal value $\boldsymbol{F}^*$ is equivalent to the solution of the following Lyapunov equation:
\begin{equation}
    \label{eq: appendix lyapunov equation}
    \begin{aligned}
        (\boldsymbol{I}-\alpha{\boldsymbol{\Bar{S}}})\boldsymbol{F} + \boldsymbol{F}(\boldsymbol{I}-\alpha{\boldsymbol{\Bar{S}}}) = 2(1-\alpha)\boldsymbol{E}.
    \end{aligned}
\end{equation}
 
\subsection{Basic Iterative Solution}
\label{appendix: basic iterative solution}

Directly solving the Lyapunov equation proposed in Eq.~\eqref{eq: appendix lyapunov equation} is time consuming, therefore, it is crucial to develop a more efficient method to approximate the solution.
In this section, we will prove that the optimal result can be infinitely approached in an iterative manner as follows:
\begin{equation}
    \label{eq: appendix iterative function}
    \boldsymbol{F}^{(t+1)} = \frac12\alpha \boldsymbol{F}^{(t)}\boldsymbol{\Bar{S}}^\top + \frac12\alpha \boldsymbol{\Bar{S}}\boldsymbol{F}^{(t)} + (1-\alpha)\boldsymbol{E}.
\end{equation}
where $\boldsymbol{S}=\boldsymbol{D}^{-1/2}\boldsymbol{W}\boldsymbol{D}^{1/2}$, and $\boldsymbol{\Bar{S}}=(\boldsymbol{S}+\boldsymbol{S}^\top)/2$ is symmetric. 
By utilizing \cref{lemma: appendix vectorization kronecker product} and incorporating the $vec(\cdot)$ operator on both sides, we can reformulate the updating process as:
\begin{equation}
    \label{eq: appendix vector iteration}
    \begin{aligned}
    vec(\boldsymbol{F}^{(t+1)}) &= \frac12\alpha(\boldsymbol{\Bar{S}}\otimes \boldsymbol{I})vec(\boldsymbol{F}^{(t)}) + \frac12\alpha(\boldsymbol{I}\otimes \boldsymbol{\Bar{S}})vec(F^{(t)}) + (1-\alpha)vec(\boldsymbol{E})  \\
    &= \alpha\Bar{\mathbb{S}}vec(\boldsymbol{F}^{(t)}) + (1-\alpha)vec(\boldsymbol{E}).
    \end{aligned}
\end{equation}
Suppose the iteration starts from an initial value of $\boldsymbol{F}^{(0)}$, for instance, $\boldsymbol{F}^{(0)}$ can be equal to matrix $\boldsymbol{E}$. 
By recursively substituting the current value into the iterative formula, we can  derive a relationship where the current value $\boldsymbol{F}^{(t)}$ depends only on the initial value $\boldsymbol{F}^{(0)}$, that is:
\begin{equation}
    vec(\boldsymbol{F}^{(t)}) = (\alpha\Bar{\mathbb{S}})^{t}vec(\boldsymbol{F}^{(0)}) + (1-\alpha)\sum_{i=0}^{t-1}(\alpha\Bar{\mathbb{S}})^ivec(\boldsymbol{E}).
\end{equation}
\begin{lemma}
    Let $\boldsymbol{A}\in \mathbb{R}^{n\times n}$, then $\lim_{k\to\infty}\boldsymbol{A}^k=0$ if and only if $\rho(\boldsymbol{A})<1$.
\end{lemma}
\begin{lemma}
    Given a matrix $\boldsymbol{A}\in\mathbb{R}^{n\times n}$ and $\rho(\boldsymbol{A})<1$, the Neumann series $\boldsymbol{I}+\boldsymbol{A}+\boldsymbol{A}^2+\cdots$ converges to $(\boldsymbol{I}-\boldsymbol{A})^{-1}$.
\end{lemma}
We have already shown that the spectral radius of $\Bar{\mathbb{S}}$ is no larger than 1, by taking advantage of these above two lemmas, we can determine the limits of the following two expressions:
\begin{equation}
    \lim_{t\to\infty}(\alpha\Bar{\mathbb{S}})^{t} = 0,
\end{equation}
and
\begin{equation}
    \lim_{t\to\infty}\sum_{i=0}^{t-1}(\alpha\Bar{\mathbb{S}})^i=(\mathbb{I}-\alpha\Bar{\mathbb{S}})^{-1}.
\end{equation}
Therefore, the iteration in Eq.~\eqref{eq: appendix vector iteration} induce to:
\begin{equation}
   vec(F^*) = (1-\alpha)(\mathbb{I}-\alpha\Bar{\mathbb{S}})^{-1}vec(E),
\end{equation}
by taking the inverse vectorization operator $vec^{-1}(\cdot)$ on both side, we can obtain that:
\begin{equation}
    \boldsymbol{F}^* = (1-\alpha)vec^{-1}\big((\mathbb{I}-\alpha\Bar{\mathbb{S}})^{-1}vec(\boldsymbol{E})\big),
\end{equation}
which is identical to the expression in Eq.~\eqref{eq: appendix optimal solution}.
Then we will analyze the convergence rate of Eq.~\eqref{eq: appendix vector iteration}. Since we have already proved that $\rho(\Bar{\mathbb{S}})\leq1$ and thus $\rho(\alpha\Bar{\mathbb{S}})<1$ given $\alpha<1$, and we can find a matrix norm $\Vert\cdot\Vert$ such that $\Vert\alpha\Bar{\mathbb{S}}\Vert<1$. More generally, with an induced matrix norm $\Vert\cdot\Vert$, the iteration follows the convergence rate:
\begin{equation}
    \Vert vec(\boldsymbol{F}^{(t)})-vec(\boldsymbol{F}^*)\Vert \leq \Vert(\alpha\Bar{\mathbb{S}})^t\Vert\cdot\Vert vec(\boldsymbol{F}^{(0)})-vec(\boldsymbol{F}^*)\Vert.
\end{equation}
\begin{lemma}
    \label{lemma: appendix matrix norm spectral radius}
    Let $\Vert\cdot\Vert$ be a matrix norm on $\mathbb{R}^{n\times n}$ and let $\boldsymbol{A} \in\mathbb{R}^{n\times n}$, then $\rho(\boldsymbol{A})=\lim_{k\to \infty}\Vert\boldsymbol{A}^k\Vert^{1/k}$.
\end{lemma}
If we define the average convergence rate at $t$-th iteration as $R_{t}(\alpha\Bar{\mathbb{S}})=-\ln{\Vert(\alpha\Bar{\mathbb{S}})^t\Vert}/t$. 
By utilizing \cref{lemma: appendix matrix norm spectral radius}, we can derive that the asymptotic convergence rate $R_{\infty}(\alpha\Bar{\mathbb{S}})$ of the error follows: 
\begin{equation}
    \label{eq: appendix basic convergence rate}
    R_{\infty}(\alpha\Bar{\mathbb{S}})=\lim_{t\to\infty}R_t(\alpha\Bar{\mathbb{S}})=-\ln\rho(\alpha\Bar{\mathbb{S}}).
\end{equation}
In addition to that, since all the entries of matrix $\boldsymbol{F}^{(0)}$, $\boldsymbol{\Bar{S}}$ and $\boldsymbol{E}$ in the iterations of Eq.~\eqref{eq: appendix vector iteration} are no less than zero, we can easily obtain that all the elements of $\boldsymbol{F}^*$ are also greater than or equal to zero. 
This characteristic is useful to our Bidirectional Similarity Diffusion process.

\subsection{Efficient Iterative Solution}
\label{appendix: efficient iterative solution}
\begin{algorithm}[t]
    \renewcommand{\algorithmicrequire}{\textbf{Input:}}
    \renewcommand{\algorithmicensure}{\textbf{Output:}}
    \caption{Efficient Iterative Solution for Bidirectional Similarity Diffusion Process}
    \label{alg: efficient iterative solution}
    \begin{algorithmic}[1]
       \REQUIRE initial estimation $\boldsymbol{F}^{(0)}\in\mathbb{R}^{n\times n}$, normalized Kronecker matrix $\Bar{\mathbb{S}}\in\mathbb{R}^{n^2\times n^2}$, identity matrix $\mathbb{I}\in\mathbb{R}^{n^2\times n^2}$, max number of iterations $maxiter$, hyper-parameter $\alpha=\frac{1}{1+\mu}$, iteration tolerance $\delta$.
       \STATE initialize $\boldsymbol{P}^{(0)}$ and $\boldsymbol{R}^{(0)}$ with $2(1-\alpha)\boldsymbol{E}-(\boldsymbol{I}-\alpha\boldsymbol{\Bar{S}})\boldsymbol{F}^{(0)} - \boldsymbol{F}^{(0)}(\boldsymbol{I}-\alpha\boldsymbol{\Bar{S}})$.
       \STATE denote $\boldsymbol{f}_t=vec(\boldsymbol{F}^{(t)})$, $\boldsymbol{r}_t=vec(\boldsymbol{R}^{(t)})$, $\boldsymbol{p}_t=vec(\boldsymbol{P}^{(t)})$.
       \FOR{$t=0,1,\dotsc, maxiter$}
       \STATE compute parameter $\alpha_t=\dfrac{\boldsymbol{r}_t^\top\boldsymbol{r}_t}{2\boldsymbol{p}_t^\top (\mathbb{I}-\alpha\Bar{\mathbb{S}})\boldsymbol{p}_t} $.
       \STATE refresh $\boldsymbol{f}_{t+1}=\boldsymbol{f}_t+\alpha_t \boldsymbol{p}_t$.
       \STATE update residue $\boldsymbol{r}_{t+1}=\boldsymbol{r}_t - 2\alpha_t(\mathbb{I}-\alpha\Bar{\mathbb{S}})\boldsymbol{p}_t$.
       \IF{$\Vert\boldsymbol{r}_{t+1}\Vert<\delta$}
           \STATE return $\boldsymbol{F}^*=vec^{-1}(\boldsymbol{f}^*)$.
       \ENDIF
       \STATE compute parameter $\beta_t=\frac{\boldsymbol{r}_{t+1}^\top\boldsymbol{r}_{t+1}}{\boldsymbol{r}_{t}^\top\boldsymbol{r}_{t}}$.
       \STATE refresh $\boldsymbol{p}_{t+1}=\boldsymbol{r}_{t+1}+\beta_t \boldsymbol{p}_t$.
       \ENDFOR
       \ENSURE $\boldsymbol{F}^*=vec^{-1}(\boldsymbol{f}^*)$.
    \end{algorithmic}
\end{algorithm}

\begin{definition}\label{conjugate_gradients_definition}
    Given a square matrix $\boldsymbol{A}\in\mathbb{R}^{n\times n}$ and a vector $\boldsymbol{x}\in\mathbb{R}^n$, the $k$-th Krylov subspace is defined as $\mathcal{K}_k(\boldsymbol{A};\boldsymbol{x})=\text{span}\{\boldsymbol{x},\boldsymbol{A}\boldsymbol{x},\boldsymbol{A}^2\boldsymbol{x},\dots,\boldsymbol{A}^{k-1}\boldsymbol{x}\}$. If $\boldsymbol{A}$ is symmetric positive definite and thus invertible, we can sort the eigenvalues of $\boldsymbol{A}$ in ascending order, \emph{i.e.}, $\lambda_1<\lambda_2<\dots<\lambda_n, \lambda_i\in\sigma(\boldsymbol{A})$, and the condition number of $\boldsymbol{A}$ is denoted as $\kappa(\boldsymbol{A})=\Vert\boldsymbol{A}\Vert_2\Vert\boldsymbol{A}^{-1}\Vert=\lambda_{n}/\lambda_{1}$. The inner product of $\boldsymbol{x},\boldsymbol{y}\in\mathbb{R}^n$ with respect to operator $\boldsymbol{A}$ is defined as $(\boldsymbol{x},\boldsymbol{y})_{\boldsymbol{A}}=\boldsymbol{A}^\top\boldsymbol{A}\boldsymbol{y}$, which can induces the $\boldsymbol{A}$-norm defined on $\mathbb{R}^n$ by $\Vert\boldsymbol{x}\Vert_{\boldsymbol{A}}=\sqrt{\boldsymbol{x}^\top\boldsymbol{A}\boldsymbol{x}}$. If $(\boldsymbol{x},\boldsymbol{y})_{\boldsymbol{A}}=0$, we say that $\boldsymbol{x}$ and $\boldsymbol{y}$ are $\boldsymbol{A}$-orthogonal.
\end{definition}
\cref{alg: efficient iterative solution} demonstrates an efficient solution to Eq.~\eqref{eq: appendix lyapunov equation} utilizing the conjugate gradient method. 
It can be viewed as minimizing the residual $\boldsymbol{r}$ over the Krylov
subspace defined by $\mathcal{K}_t(\mathbb{I}-\alpha\Bar{\mathbb{S}};\boldsymbol{r}_0)$ in an iterative way. 
The iteration starts from an initial value $\boldsymbol{F}^{(0)}$, and denote the result after $t$-th iteration by $\boldsymbol{F}^{(t)}$. 
Take the vectorize operation $vec(\cdot)$ during the updating process, the error is measured by the $(\mathbb{I}-\alpha\Bar{\mathbb{S}})$-norm of the difference between $vec(\boldsymbol{F}^{(t)})$ and the optimum value $vec(\boldsymbol{F}^*)$, with the following convergence rate:
\begin{equation}
    \Vert vec(\boldsymbol{F}^{(t)})-vec(\boldsymbol{F}^*)\Vert^2_{\mathbb{I}-\alpha\Bar{\mathbb{S}}} \leq 
\inf_{\substack{p_t\in\mathcal{P}_t \\p_t(0)=1}}\sup_{\lambda}|p_t(\lambda)|^2\Vert vec(\boldsymbol{F}^{(0)})-vec(\boldsymbol{F}^*)\Vert^2_{\mathbb{I}-\alpha\Bar{\mathbb{S}}},
\end{equation}
where $p_t(x)\in\mathcal{P}_t$ is the so called residual polynomial constrained by $p_t(0)=1$. It guarantees that the optimal solution will be found within a limited number of iterations even in the worst case. 
By selecting specific residual polynomials, we can derive various upper bounds for convergence.
In particular, we can utilize the following Chebyshev polynomial $T_t(x)$ to define the residual polynomial $p_t(x)$:
\begin{equation}
    T_t(x)=\left\{ 
    \begin{array}{lc}
        \cos(t\cdot\arccos{x}), & |x| \leqslant 1 \\
        \cosh(t\cdot\text{arccosh}{x}), & |x|> 1\\
    \end{array}
    \right.
\end{equation}
and thus the convergence rate can be obtained by:
\begin{equation}
    \Vert vec\boldsymbol{F}^{(t)})-vec(\boldsymbol{F}^*)\Vert_{\mathbb{I}-\alpha\Bar{\mathbb{S}}} \leq 2\Big(\frac{\sqrt{\kappa(\mathbb{I}-\alpha\Bar{\mathbb{S}})}-1}{\sqrt{\kappa(\mathbb{I}-\alpha\Bar{\mathbb{S}})}+1}\Big)^t\Vert vec(\boldsymbol{F}^{(0)})-vec(\boldsymbol{F}^*)\Vert_{\mathbb{I}-\alpha\Bar{\mathbb{S}}},
\end{equation}
where $\kappa(\mathbb{I}-\alpha\Bar{\mathbb{S}})$ is the corresponding condition number mentioned in \cref{conjugate_gradients_definition}. The convergence will be even faster if the eigenvalues are clustered. Assume $\sigma(\mathbb{I}-\alpha\Bar{\mathbb{S}})=\sigma_0\bigcup\sigma_1$, and the number of elements in $\sigma_0$ is $l$, we can obtain a corollary shown in Eq.~\eqref{eq:convergence_corollary_conjugate_gradients}, which implies that the convergence rate is mainly governed by the effective condition number $b/a$,
\begin{equation}\label{eq:convergence_corollary_conjugate_gradients}
    \Vert vec(\boldsymbol{F}^{(t)})-vec(\boldsymbol{F}^*)\Vert \leq 2M\Big(\frac{\sqrt{b/a}-1}{\sqrt{b/a}+1}\Big)^{t-l}\Vert vec(\boldsymbol{F}^{(0)})-vec(\boldsymbol{F}^*)\Vert,
\end{equation}
in which:
\begin{equation*}
    a=\min_{\lambda\in\sigma_1}\lambda,\text{ } b=\max_{\lambda\in\sigma_1}\lambda,\text{ } \text{and } M=\max_{\lambda\in\sigma_1}\prod_{\mu\in\sigma_0}\vert 1-\lambda/\mu\vert.
\end{equation*}

\section{Neighbor-guided Similarity Smooth}
\label{appendix: feature enhancement}
In this section, we will first derive a more generalized formulation of our proposed optimization problem as below.
The goal of this proposition is to find an optimal vector $\boldsymbol{x}$ to minimize the quadratic objective function, note that all the elements of $\boldsymbol{f}$ and $\boldsymbol{p}$ are no less than 0, and all the entries of $\boldsymbol{p}$ are also not exceed $r$, \emph{i.e.}, $0\leq\boldsymbol{p}_i\leq r$ for $i=1,2\dots,n$.
\begin{equation}
    \begin{aligned}
        \mbox{minimize} \quad& \frac12\Vert r\boldsymbol{x}-\boldsymbol{p}\circ\boldsymbol{f}\Vert^2_2 + \beta\Vert\boldsymbol{x}-\boldsymbol{f}\Vert^2_2 \\
        \mbox{subject to}\quad & \boldsymbol{x}_i\geq0,\quad i=1,2\dots,n \\
                    & \Vert\boldsymbol{x}\Vert_1=\Vert\boldsymbol{f}\Vert_1,
    \end{aligned}
\end{equation}
where $\beta>0$ is the regularization weight, and symbol $\circ$ is an element-wise multiplier, capable of merging two vectors.
We first rewrite the optimization problem into a standard form. Note that both the objective function and constraints are convex, thus the primal problem is also convex,
\begin{equation}
    \begin{aligned}
        \mbox{minimize} \quad& \frac12\Vert r\boldsymbol{x}-\boldsymbol{p}\circ\boldsymbol{f}\Vert^2_2 + \beta\Vert\boldsymbol{x}-\boldsymbol{f}\Vert^2_2 \\
        \mbox{subject to}\quad &-\boldsymbol{x}_i\leq0,\quad i=1,2\dots,n \\
        &\sum_{i=1}^n \boldsymbol{x}_i - \sum_{i=1}^n \boldsymbol{f}_i = 0.
    \end{aligned}
\end{equation}
Denote $\mathcal{D}$ as the domain defined by the inequality constraints. Since there exists a feasible point $\boldsymbol{x}^*\in \textbf{refine\ }\mathcal{D}$ that makes the inequality constraints strictly hold. 
Thus the convex optimization problem satisfies Slater’s constraint qualification, which states that the strong duality holds and the KKT conditions can provide sufficiency and necessity for finding the optimal solution. 
The Lagrange function $\mathcal{L}(\boldsymbol{x},\boldsymbol{\lambda},\nu)$ corresponding to the primal constrained optimization problem can formulated as below:
\begin{equation}
    \begin{aligned}
        \mathcal{L}(\boldsymbol{x},\boldsymbol{\lambda},\nu)=\frac12\Vert r\boldsymbol{x}-\boldsymbol{p}\circ\boldsymbol{f}\Vert^2_2 + \beta\Vert\boldsymbol{x}-\boldsymbol{f}\Vert^2_2-\sum_{i=1}^{n}\boldsymbol{\lambda}_i\boldsymbol{x}_i + \nu(\sum_{i=1}^n \boldsymbol{x}_i - \sum_{i=1}^n \boldsymbol{f}_i).
    \end{aligned} 
\end{equation}
Suppose $\widetilde{\boldsymbol{x}}$, $\boldsymbol{\widetilde{\lambda}}$, $\widetilde{\nu}$ satisfy the following KKT conditions. Since the Slater's condition holds for this convex optimization problem, $\widetilde{\boldsymbol{x}}$ and $\boldsymbol{\widetilde{\lambda}},\widetilde{\nu}$ are primal and dual optimal with zero duality gap and the optimum is attained,
\begin{equation}
    \label{eq:KKT}
    \begin{aligned}
        -\widetilde{\boldsymbol{x}}_i&\leq0,\ i=1,\dots,n \\
        \sum_{i=1}^n\widetilde{\boldsymbol{x}}_i-\sum_{i=1}^n \boldsymbol{f}_i&=0\\
        \boldsymbol{\widetilde{\lambda}}_i &\geq0,\ i=1,\dots,n \\
        \boldsymbol{\widetilde{\lambda}}_i\widetilde{\boldsymbol{x}}_i&=0,\ i=1,\dots,n\\
        r^2\widetilde{\boldsymbol{x}}_i-r\boldsymbol{p}_i\boldsymbol{f}_i + 2\beta(\widetilde{\boldsymbol{x}}_i-\boldsymbol{f}_i) - \widetilde{\boldsymbol{\lambda}}_i + \widetilde{\nu} &= 0,\ i=1,\dots,n.
    \end{aligned}
\end{equation}
First, we prove that $\boldsymbol{\widetilde{\lambda}}_i=0$ for $i=1,2,\dots,n$, then we can get the closed-form solution to the optimization problem more easily.
Suppose there exists some $i$ such that $\boldsymbol{\widetilde{\lambda}}_{i}>0$ and thus the corresponding $\widetilde{\boldsymbol{x}}_{i}=0$ according to the complementary slackness condition.
We can obtain that $\widetilde{\boldsymbol{\lambda}}_{i}=\widetilde{\nu}-r\boldsymbol{p}_{i}\boldsymbol{f}_{i}-2\beta \boldsymbol{f}_i$. And it is worth to notice that not all the $\widetilde{\boldsymbol{x}}_i=0$ since $\Vert\widetilde{\boldsymbol{x}}\Vert_1=\Vert\boldsymbol{f}\Vert_1$. Taking the sum of all the $\widetilde{\boldsymbol{\lambda}}_i$ by using the last KKT condition in Eq.~\eqref{eq:KKT}, we can get:
\begin{equation}
    \label{lambda_summation1}
    \sum_{i=1}^n \widetilde{\boldsymbol{\lambda}}_i= n\widetilde{\nu} + r^2\sum_{i=1}^n \boldsymbol{f}_i - r\sum_{i=1}^n \boldsymbol{p}_i\boldsymbol{f}_i,
\end{equation}
which is equal to the summation of all the $\widetilde{\boldsymbol{\lambda}}_{i}\neq0$, that is:
\begin{equation}
    \label{lambda_summation2}
    \sum_{i,\widetilde{\boldsymbol{\lambda}}_{i}\neq0} \widetilde{\boldsymbol{\lambda}}_{i} = \sum_{i,\widetilde{\boldsymbol{\lambda}}_{i}\neq0}\widetilde{\nu} - r\boldsymbol{p}_{i}\boldsymbol{f}_{i} - 2\beta \boldsymbol{f}_i.
\end{equation}
Take some small transformations to the two Eq.~\eqref{lambda_summation1} and Eq.~\eqref{lambda_summation2}, we can derive that:
\begin{equation*}
    \sum_{i,\widetilde{\boldsymbol{\lambda}}_{i}=0}\widetilde{\nu} = r\sum_{i=1}^n (\boldsymbol{p}_i\boldsymbol{f}_i-r\boldsymbol{f}_i) -  \sum_{i,\widetilde{\boldsymbol{\lambda}}_{i}\neq0}(r\boldsymbol{p}_{i}\boldsymbol{f}_{i} + 2\beta \boldsymbol{f}_i).
\end{equation*}
As we have $\boldsymbol{p}_i\leq r$ in our settings, we can infer that $\widetilde{\nu}\leq0$, which will lead to a contradiction that $\widetilde{\boldsymbol{\lambda}}_{i}=\widetilde{\nu}-r\boldsymbol{p}_{i}\boldsymbol{f}_{i}-2\beta \boldsymbol{f}_i\leq0$ for all $i$ when we are expecting there exists some $\widetilde{\boldsymbol{\lambda}}_{i}>0$. Thus, we have proved that $\widetilde{\boldsymbol{\lambda}}_i=0$ for $i=1,2\dots,n$. From Eq.~\eqref{lambda_summation1}, we can solve the optimal value of $\widetilde{\nu}$ as:
\begin{equation}
    \widetilde{\nu} = r\sum_{i=1}^n (\boldsymbol{p}_i\boldsymbol{f}_i-r\boldsymbol{f}_i)/n.
\end{equation}
Bring back $\widetilde{\nu}$ into the last equation in the KKT conditions, we can find the optimal value as below:
\begin{equation}
    \widetilde{\boldsymbol{x}}_i = \frac{r\boldsymbol{p}_i+2\beta}{r^2+2\beta}\boldsymbol{f}_i + \frac{r\sum_{i=1}^n(r\boldsymbol{f}_i-\boldsymbol{p}_i\boldsymbol{f}_i)}{n(r^2+2\beta)},
\end{equation}
which can also be expressed in vector form as:
\begin{equation}
    \widetilde{\boldsymbol{x}} = \frac{r\boldsymbol{p}+2\beta}{r^2+2\beta}\circ\boldsymbol{f} + \frac{r^2\Vert\boldsymbol{f}\Vert_1-r\boldsymbol{p}^\top\boldsymbol{f}}{n(r^2+2\beta)}.
\end{equation}
In addition, we introduce a sparse variation to the original optimization problem. If $\boldsymbol{f}$ is a sparse vector with activation dimension set $\mathcal{C}$, and the optimized vector $\boldsymbol{x}$ share the same sparsity, the objective function then becomes:
\begin{equation}
    \begin{aligned}
        \mbox{minimize} \quad& \frac12\Vert r\boldsymbol{x}-\boldsymbol{p}\circ\boldsymbol{f}\Vert^2_2 + \beta\Vert\boldsymbol{x}-\boldsymbol{f}\Vert^2_2 \\
        \mbox{subject to}\quad & \boldsymbol{x}_i\geq0,\quad i\in\mathcal{C} \\
        & \boldsymbol{x}_i=0, \quad i=1,2\dots,\mathcal{|X|}/\mathcal{C}\\
                    & \Vert\boldsymbol{x}\Vert_1=\Vert\boldsymbol{f}\Vert_1.
    \end{aligned}
\end{equation}
It can be viewed as a special case of the general optimization problem, the optimal solution can be quickly obtained by:
\begin{equation}
    \widetilde{\boldsymbol{x}}_i = \frac{r\boldsymbol{p}_i+2\beta}{r^2+2\beta}\boldsymbol{f}_i + \frac{r\sum_{i\in\mathcal{C}}(r\boldsymbol{f}_i-\boldsymbol{p}_i\boldsymbol{f}_i)}{|\mathcal{C}|(r^2+2\beta)}, \quad i\in\mathcal{C}.
\end{equation}
As for the proposed Neighbor-guided Similarity Smooth problem proposed in \cref{Neighbor-guided Similarity Smooth}. We can leverage the above formula to get the closed-form solution as below:
\begin{equation}
    \boldsymbol{\hat{F}}_{ij} = \frac{r\boldsymbol{T}_{ij}+2\beta}{r^2+2\beta}\boldsymbol{F}_{ij} + \frac{r^2\Vert\boldsymbol{F}_i\Vert_1-r\boldsymbol{T}_i^\top\boldsymbol{F}_i}{\vert\mathcal{C}[i]\vert(r^2+2\beta)}, \quad j\in\mathcal{C}[i].
\end{equation}

\begin{table*}[t]
    \caption{Evaluation of the performance on Market1501. Here we select BoT \cite{cvprw2019_bot}, OSNet \cite{tpami2022_osnet}, AGW \cite{tpami2022_agw}, MGN \cite{mm2018_mgn} and SpCL \cite{nips2020_spcl} as our backbone model.}
    \label{tab:appendix market1}
    \vskip 0.15in
    \centering
    \resizebox{0.96\textwidth}{!}{
    \begin{tabular}{cccccccccccccccc}
    \toprule 
    \multicolumn{1}{c}{\multirow{2.4}*{Method}} & \multicolumn{3}{c}{BoT} & \multicolumn{3}{c}{OSNet} & \multicolumn{3}{c}{AGW} & \multicolumn{3}{c}{MGN} & \multicolumn{3}{c}{SpCL}  \\
    \cmidrule(lr){2-4}
    \cmidrule(lr){5-7}
    \cmidrule(lr){8-10}
    \cmidrule(lr){11-13}
    \cmidrule(lr){14-16}
    ~ & mAP & R1 & mINP & mAP & R1 & mINP & mAP & R1 & mINP & mAP & R1 & mINP & mAP & R1 & mINP  \\
    \midrule  
    Baseline & 86.4 & 94.7 & 60.9 & 85.8 & 94.8 & 57.1 & 88.4 & 95.6 & 66.1 & 89.9 & 95.8 & 68.1 & 72.2 & 87.7 & 33.3 \\
    kNN & 88.8 & 95.2 & 65.3 & 89.1 & 95.5 & 66.5 & 90.1 & 96.0 & 68.3 & 91.7 & 96.4 & 72.4 & 77.5 & 89.3 & 42.3 \\
    AQE & 93.3 & 95.4 & 85.6 & 92.6 & 95.0 & 84.1 & 93.9 & 95.8 & 87.3 & 94.8 & 96.4 & 89.3 & 82.8 & 89.6 & 58.5 \\
    $\alpha$QE & 92.9 & 95.6 & 82.8 & 91.4 & 95.0 & 81.5 & 93.6 & 96.0 & 84.9 & 94.8 & 96.5 & 87.4 & 78.4 & 88.6 & 50.7 \\
    AQEwD & 93.1 & 95.3 & 84.8 & 92.5 & 94.8 & 83.6 & 93.7 & 95.8 & 86.6 & 94.6 & 96.4 & 88.6 & 83.6 & 90.3 & 57.3 \\
    RDP & 93.6 & 95.8 & 87.5 & 93.2 & 95.7 & 86.0 & 93.9 & 96.2 & 88.5 & 95.1 & 96.8 & 90.7 & 85.7 & 90.4 & 64.3 \\
    SCA & 93.8 & 94.9 & 88.5 & 93.3 & 95.0 & 86.4 & 94.1 & 95.4 & 89.3 & 95.3 & 96.3 & 91.2 & 85.7 & 90.0 & 64.7 \\
    $k$-recip & 94.2 & 95.6 & 88.6 & 93.7 & 95.2 & 86.8 & 94.7 & 95.9 & 89.5 & 95.4 & \textbf{97.0} & 90.1 & 86.1 & 90.6 & 63.3 \\
    ECN & 94.0 & 95.6 & 88.6 & 93.7 & 95.5 & 87.0 & 94.3 & 96.0 & 89.5 & 95.5 & 96.7 & 91.5 & 86.2 & 90.7 & 65.5 \\
    \midrule
    Ours & \textbf{95.2} & \textbf{96.1} & \textbf{91.2} & \textbf{94.6} & \textbf{95.7} & \textbf{88.8} & \textbf{95.2} & \textbf{96.4} & \textbf{91.1} & \textbf{95.9} & 96.7 & \textbf{92.3} & \textbf{87.8} & \textbf{91.5} & \textbf{67.7} \\
    \bottomrule
    \end{tabular}}
\end{table*}

\section{Sparse Jensen-Shannon Divergence}
\label{section: appendix sparse jensen shannon divergence}

Instead of directly using the optimized similarity matrix $\boldsymbol{F}$ for instance retrieval, we apply Eq.~\eqref{eq: sd} to compute the Jensen-Shannon divergence as our distance measure.
Note that each row in $\boldsymbol{F}$ is post-processed with an $l_1$-normalization operation, which can be viewed as a probability distribution within the data manifold. This approach aligns more closely with our similarity diffusion process.
Recall the divergence function as below:
\begin{equation}
    \label{eq: appendix sparse jensen shannon divergence}
    \begin{aligned}
    d_{JS}(i,j) = \frac{1}{2}\sum_{k=1}^{n}\boldsymbol{F}_{ik}\log\Big(\frac{2\boldsymbol{F}_{ik}}{\boldsymbol{F}_{ik}+\boldsymbol{F}_{jk}}\Big) + \frac{1}{2}\sum_{k=1}^{n}\boldsymbol{F}_{jk}\log\Big(\frac{2\boldsymbol{F}_{jk}}{\boldsymbol{F}_{ik}+\boldsymbol{F}_{jk}}\Big).
    \end{aligned}   
\end{equation}
The computation speed can be enhanced by leveraging the sparsity of the similarity matrix.
For the first part in Eq.~\eqref{eq: appendix sparse jensen shannon divergence}, we can decompose and rewrite this term as follows:
\begin{equation}
    \begin{aligned}
    \frac{1}{2}\sum_{k=1}^{n}\boldsymbol{F}_{ik}&\log\Big(\frac{2\boldsymbol{F}_{ik}}{\boldsymbol{F}_{ik}+\boldsymbol{F}_{jk}}\Big) = 
    \frac{1}{2}\sum_{k|\boldsymbol{F}_{ik}\neq0}\boldsymbol{F}_{ik}\log\Big(\frac{2\boldsymbol{F}_{ik}}{\boldsymbol{F}_{ik}+\boldsymbol{F}_{jk}}\Big) +& \frac{1}{2}\sum_{k|\boldsymbol{F}_{ik}=0}\boldsymbol{F}_{ik}\log\Big(\frac{2\boldsymbol{F}_{ik}}{\boldsymbol{F}_{ik}+\boldsymbol{F}_{jk}}\Big),
    \end{aligned}
\end{equation}
where the corresponding item when $\boldsymbol{F}_{ik}=0$ is obviously equal to zero.
As for the situation when $\boldsymbol{F}_{ik}=0$, we can make some further decomposition depending on whether $\boldsymbol{F}_{jk}$ is equal to 0, that is:
\begin{equation}
    \begin{aligned}
    \frac{1}{2}\sum_{k|\boldsymbol{F}_{ik}\neq0}\boldsymbol{F}_{ik}\log\Big(\frac{2\boldsymbol{F}_{ik}}{\boldsymbol{F}_{ik}+\boldsymbol{F}_{jk}}\Big) 
    =\frac{1}{2}\sum_{k|\boldsymbol{F}_{ik}\neq0, \boldsymbol{F}_{jk}=0}\boldsymbol{F}_{ik}\log\Big(\frac{2\boldsymbol{F}_{ik}}{\boldsymbol{F}_{ik}+\boldsymbol{F}_{jk}}\Big) +\frac{1}{2}\sum_{k|\boldsymbol{F}_{ik}\neq0, \boldsymbol{F}_{jk}\neq0}\boldsymbol{F}_{ik}\log\Big(\frac{2\boldsymbol{F}_{ik}}{\boldsymbol{F}_{ik}+\boldsymbol{F}_{jk}}\Big). \\
    \end{aligned}
\end{equation}
Based on the observation that the log function in the expression is equal to 1 while $\boldsymbol{F}_{ik}\neq0$ and $\boldsymbol{F}_{jk}=0$, and by leveraging the fact that the $l_1$-norm of $\boldsymbol{F}_i$ equal to 1, \emph{i.e.}, $\Vert\boldsymbol{F}_i\Vert_1=1$, we can obtain that:
\begin{equation}
    \begin{aligned}
    \frac{1}{2}\sum_{k|\boldsymbol{F}_{ik}\neq0}\boldsymbol{F}_{ik}\log\Big(\frac{2\boldsymbol{F}_{ik}}{\boldsymbol{F}_{ik}+\boldsymbol{F}_{jk}}\Big) &=  \frac{1}{2}\sum_{k|\boldsymbol{F}_{ik}\neq0}\boldsymbol{F}_{ik} -  \frac{1}{2}\sum_{k|\boldsymbol{F}_{ik}\neq0, \boldsymbol{F}_{jk}\neq0}\boldsymbol{F}_{ik} + \frac{1}{2}\sum_{k|\boldsymbol{F}_{ik}\neq0, \boldsymbol{F}_{jk}\neq0}\boldsymbol{F}_{ik}\log\Big(\frac{2\boldsymbol{F}_{ik}}{\boldsymbol{F}_{ik}+\boldsymbol{F}_{jk}}\Big)\\
    &= \frac{1}{2} + \frac{1}{2}\sum_{k|\boldsymbol{F}_{ik}\neq0, \boldsymbol{F}_{jk}\neq0}\boldsymbol{F}_{ik}\bigg(\log\Big(\frac{2\boldsymbol{F}_{ik}}{\boldsymbol{F}_{ik}+\boldsymbol{F}_{jk}}\Big)-1\bigg).
    \end{aligned}
\end{equation}
Similar decomposition can be conducted to the second term in Eq.~\eqref{eq: appendix sparse jensen shannon divergence}, substituting them back into the original formula, the final Jensen-Shannon divergence can be efficiently computed by:
\begin{equation}
    \begin{aligned}
     d_{JS}&(i,j) = 
     1 +  \frac{1}{2}\sum_{k|\boldsymbol{F}_{ik}\neq0, \boldsymbol{F}_{jk}\neq0}\boldsymbol{F}_{ik}\bigg(\log\Big(\frac{2\boldsymbol{F}_{ik}}{\boldsymbol{F}_{ik}+\boldsymbol{F}_{jk}}\Big)-1\bigg)
     +  \frac{1}{2}\sum_{k|\boldsymbol{F}_{ik}\neq0, \boldsymbol{F}_{jk}\neq0}\boldsymbol{F}_{jk}\bigg(\log\Big(\frac{2\boldsymbol{F}_{jk}}{\boldsymbol{F}_{ik}+\boldsymbol{F}_{jk}}\Big)-1\bigg).
    \end{aligned}
\end{equation}

\section{Extended Experiments}
\label{section: appendix extended results}
\begin{figure}[ht]
    \centering
    \includegraphics[width=0.96\linewidth]{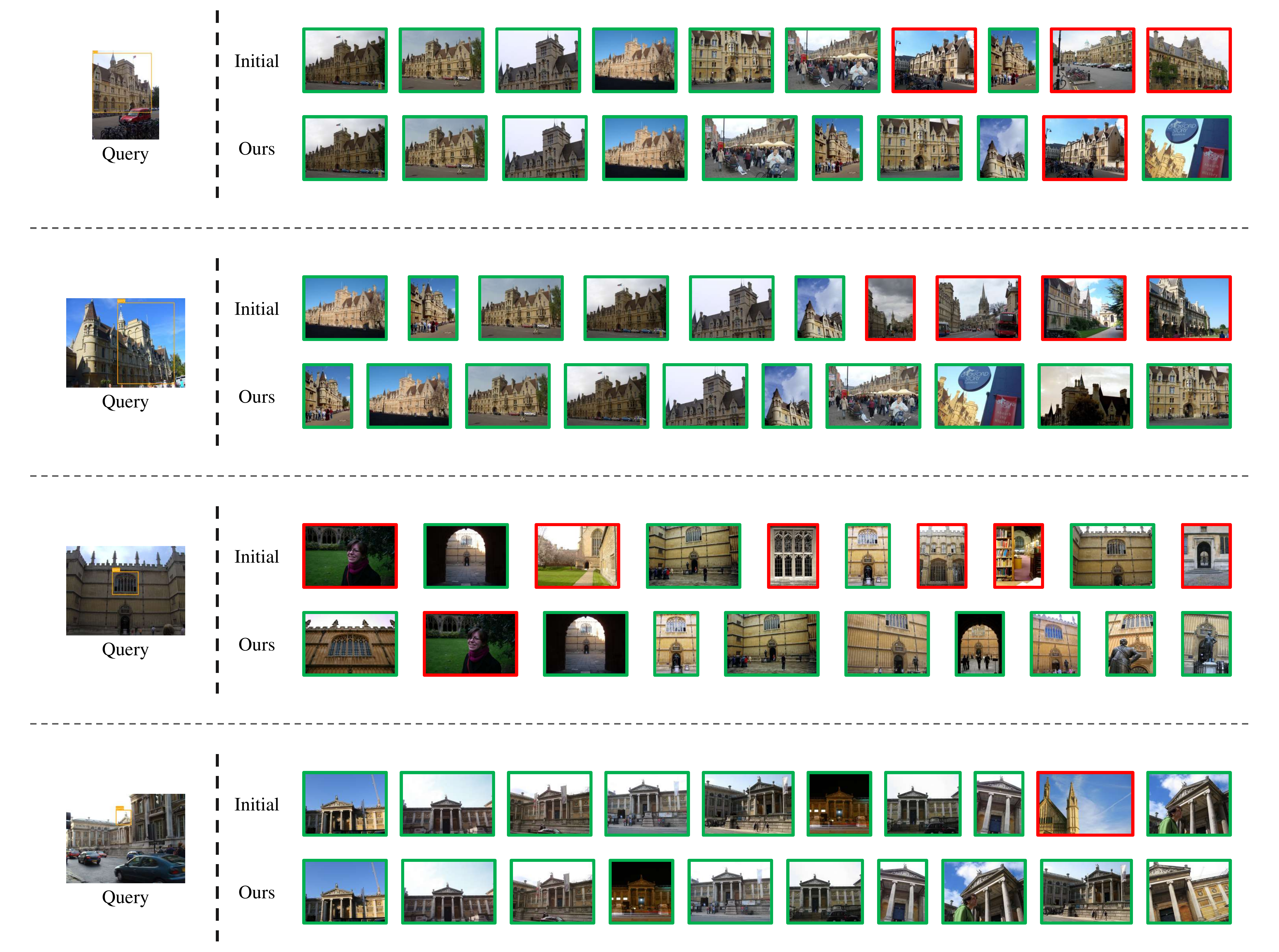}
    \caption{We showcase the retrieval performance of the initial search and our proposed method through selected qualitative examples, tested on the \emph{R}Oxford dataset. The interest region in the images on the left side are extracted with an orange bounding box as the query. On the right side, the top 10 retrieval results from both the initial search and our proposed method are presented, with true positives marked by a green bounding box and false matches denoted by a red bounding box.}
    \label{fig: appendix showcase}
\end{figure}

\begin{table*}[ht]
    \caption{Evaluation of the performance on Market1501. Here we select AdaSP \cite{cvpr2023_adasp_reid}, ABD-Net \cite{iccv2019_abd_net}, TransReID \cite{iccv2021_trans_reid} and ISE \cite{cvpr2022_ise_reid}.}
    \label{tab: appendix market2}
    \vskip 0.15in
    \centering
    \resizebox{0.96\textwidth}{!}{
    \begin{tabular}{cccccccccccccccc}
    \toprule 
    \multicolumn{1}{c}{\multirow{2.4}*{Method}} & \multicolumn{3}{c}{AdaSP} & \multicolumn{3}{c}{ABD-Net} & \multicolumn{3}{c}{TransReID} & \multicolumn{3}{c}{CLIP-ReID} & \multicolumn{3}{c}{ISE}  \\
    \cmidrule(lr){2-4}
    \cmidrule(lr){5-7}
    \cmidrule(lr){8-10}
    \cmidrule(lr){11-13}
    \cmidrule(lr){14-16}
    ~ & mAP & R1 & mINP & mAP & R1 & mINP & mAP & R1 & mINP & mAP & R1 & mINP & mAP & R1 & mINP  \\
    \midrule  
    Baseline & 86.7 & 93.9 & 62.5 & 88.5 & 95.4 & 65.4 & 88.2 & 94.7 & 67.2 & 89.8 & 95.5 & 69.3 & 83.9 & 93.0 & 54.8 \\
    kNN & 88.7 & 93.8 & 70.9 & 91.3 & 95.7 & 74.5 & 90.8 & 95.3 & 75.2 & 92.3 & 96.1 & 77.8 & 87.5 & 93.9 & 64.5 \\
    AQE & 92.2 & 94.4 & 83.1 & 93.5 & 95.8 & 87.4 & 92.0 & 94.9 & 85.3 & 94.4 & 95.9 & 89.4 & 90.3 & 93.6 & 79.8 \\
    $\alpha$QE & 92.5 & 94.6 & 83.2 & 92.1 & 95.3 & 84.1 & 90.9 & 94.5 & 82.8 & 94.4 & 96.0 & 89.1 & 86.4 & 93.1 & 71.9 \\
    AQEwD & 92.2 & 94.6 & 82.6 & 93.6 & 95.8 & 86.3 & 92.2 & 94.6 & 84.6 & 94.3 & 96.2 & 88.6 & 91.2 & 94.4 & 79.2 \\
    RDP & 92.9 & 94.8 & 85.8 & 94.3 & 96.1 & 89.3 & 93.3 & \textbf{95.9} & 87.1 & 94.9 & 96.2 & 90.3 & 92.2 & 94.7 & 83.2 \\
    SCA & 92.6 & 94.2 & 85.0 & 94.4 & 95.7 & 89.6 & 93.0 & 94.3 & 87.4 & 94.9 & 95.7 & 90.8 & 92.2 & 94.3 & 83.6 \\
    $k$-recip & 93.0 & 94.5 & 85.6 & 94.7 & \textbf{96.4} & 89.0 & 93.6 & 95.0 & 87.3 & 95.0 & 96.1 & 90.1 & 92.3 & 94.6 & 81.3 \\
    ECN & 93.0 & 94.9 & 85.8 & 94.7 & 96.3 & 90.2 & 93.8 & 95.8 & 88.0 & 95.2 & 96.5 & 91.2 & 92.4 & 94.7 & 84.1 \\
    \midrule     
    Ours & \textbf{93.7} & \textbf{95.0} & \textbf{87.5} & \textbf{95.0} & 96.2 & \textbf{90.6} & \textbf{94.6} & 95.7 & \textbf{89.7} & \textbf{95.4} & \textbf{96.5} & \textbf{91.6} & \textbf{93.4} & \textbf{95.0} & \textbf{84.9} \\
    \bottomrule
    \end{tabular}}
\end{table*}

\begin{table}
    \label{tab: appendix cuhk03}
    \caption{Evaluation of the performance on CUHK03-D and CUHK03-L. We select BoT \cite{cvprw2019_bot}, AGW \cite{tpami2022_agw}, MGN \cite{mm2018_mgn} and AdaSP \cite{cvpr2023_adasp_reid} as our backbone model.}
    \vskip 0.1in
    \centering
    \begin{minipage}{0.48\textwidth}
        \centering
        \resizebox{0.92\textwidth}{!}{
        \begin{tabular}{ccccccc}
        \toprule
        \makebox[0.23\textwidth]{\multirow{2.4}{*}{Method}} & \multicolumn{3}{c}{CUHK03-D} & \multicolumn{3}{c}{CUHK03-L}\\
        \cmidrule(lr){2-4}
        \cmidrule(lr){5-7}
        ~ & \makebox[0.12\textwidth][c]{mAP} & \makebox[0.12\textwidth][c]{R1} & \makebox[0.12\textwidth][c]{mINP} & \makebox[0.12\textwidth][c]{mAP} & \makebox[0.12\textwidth][c]{R1} & \makebox[0.12\textwidth][c]{mINP} \\
        \midrule
        BoT & 63.0 & 64.4 & 51.6 & 66.1 & 67.0 & 54.8 \\
        \midrule
        kNN & 67.3 & 68.5 & 57.5 & 70.2 & 71.0 & 60.4 \\
        AQE & 75.5 & 73.6 & 71.5 & 77.8 & 75.6 & 74.8 \\
        $\alpha$QE & 74.8 & 73.4 & 70.0 & 77.2 & 75.1 & 73.5 \\
        RDP & 78.3 & 77.1 & 75.6 & 79.7 & 77.9 & 77.4 \\
        SCA & 80.7 & 75.8 & 79.1 & 82.9 & 77.8 & 81.7 \\
        $k$-recip & 79.6 & 76.2 & 78.3 & 81.6 & 78.3 & 81.1 \\
        ECN & 80.5 & 77.5 & 79.6 & 82.1 & 79.0 & 81.8 \\
        \midrule
        Ours & \textbf{81.6} & \textbf{78.7} & \textbf{81.4} & \textbf{83.7} & \textbf{80.6} & \textbf{83.7} \\
        \bottomrule 
        \end{tabular}}
    \end{minipage}\begin{minipage}{0.48\textwidth}
        \centering
        \resizebox{0.92\textwidth}{!}{
        \begin{tabular}{ccccccc}
        \toprule
        \makebox[0.23\textwidth]{\multirow{2.4}{*}{Method}} & \multicolumn{3}{c}{CUHK03-D} & \multicolumn{3}{c}{CUHK03-L}\\
        \cmidrule(lr){2-4}
        \cmidrule(lr){5-7}
        ~ & \makebox[0.12\textwidth][c]{mAP} & \makebox[0.12\textwidth][c]{R1} & \makebox[0.12\textwidth][c]{mINP} & \makebox[0.12\textwidth][c]{mAP} & \makebox[0.12\textwidth][c]{R1} & \makebox[0.12\textwidth][c]{mINP} \\
        \midrule
        AGW & 69.8 & 71.4 & 59.4 & 72.6 & 73.4 & 62.7 \\
        \midrule
        kNN & 73.8 & 74.5 & 65.4 & 76.8 & 78.1 & 69.3 \\
        AQE & 80.0 & 78.5 & 77.4 & 83.1 & 81.9 & 80.7 \\
        $\alpha$QE & 79.1 & 77.7 & 75.6 & 82.1 & 80.6 & 78.8 \\
        RDP & 82.5 & 80.1 & 80.8 & 85.7 & 84.4 & 84.0 \\
        SCA & 84.1 & 79.5 & 83.0 & 87.1 & 83.1 & 86.2 \\
        $k$-recip & 84.0 & 81.2 & 83.2 & 86.4 & 83.5 & 85.9 \\
        ECN & 83.7 & 80.4 & 83.4 & 86.2 & 83.3 & 86.0 \\
        \midrule
        Ours & \textbf{85.8} & \textbf{82.9} & \textbf{85.5} & \textbf{87.7} & \textbf{85.2} & \textbf{87.9} \\
        \bottomrule 
        \end{tabular}}
    \end{minipage}
    \vskip 0.15in
    \centering
    \begin{minipage}{0.48\textwidth}
        \centering
        \resizebox{0.92\textwidth}{!}{
        \begin{tabular}{ccccccc}
        \toprule
        \makebox[0.23\textwidth]{\multirow{2.4}{*}{Method}} & \multicolumn{3}{c}{CUHK03-D} & \multicolumn{3}{c}{CUHK03-L}\\
        \cmidrule(lr){2-4}
        \cmidrule(lr){5-7}
        ~ & \makebox[0.12\textwidth][c]{mAP} & \makebox[0.12\textwidth][c]{R1} & \makebox[0.12\textwidth][c]{mINP} & \makebox[0.12\textwidth][c]{mAP} & \makebox[0.12\textwidth][c]{R1} & \makebox[0.12\textwidth][c]{mINP} \\
        \midrule
        MGN & 72.2 & 74.6 & 61.3 & 75.9 & 77.9 & 65.9 \\
        \midrule
        kNN & 76.2 & 77.1 & 67.7 & 80.2 & 80.9 & 72.7 \\
        AQE & 83.3 & 81.8 & 80.3 & 86.3 & 84.6 & 84.3 \\
        $\alpha$QE & 82.0 & 80.7 & 78.3 & 85.5 & 84.2 & 82.8 \\
        RDP & 86.1 & 85.1 & 83.7 & 88.3 & 87.1 & 86.8 \\
        SCA & 87.7 & 84.2 & 86.5 & 90.5 & 87.4 & 89.6 \\
        $k$-recip & 87.1 & 84.8 & 86.4 & 89.6 & 87.9 & 89.2 \\
        ECN & 86.9 & 84.6 & 86.3 & 89.8 & 87.9 & 89.5 \\
        \midrule
        Ours & \textbf{88.6} & \textbf{87.1} & \textbf{88.1} & \textbf{90.6} & \textbf{89.0} & \textbf{90.5} \\
        \bottomrule 
        \end{tabular}}
    \end{minipage}\begin{minipage}{0.48\textwidth}
        \centering
        \resizebox{0.92\textwidth}{!}{
        \begin{tabular}{ccccccc}
        \toprule
        \makebox[0.23\textwidth]{\multirow{2.4}{*}{Method}} & \multicolumn{3}{c}{CUHK03-D} & \multicolumn{3}{c}{CUHK03-L}\\
        \cmidrule(lr){2-4}
        \cmidrule(lr){5-7}
        ~ & \makebox[0.12\textwidth][c]{mAP} & \makebox[0.12\textwidth][c]{R1} & \makebox[0.12\textwidth][c]{mINP} & \makebox[0.12\textwidth][c]{mAP} & \makebox[0.12\textwidth][c]{R1} & \makebox[0.12\textwidth][c]{mINP} \\
        \midrule
        AdaSP & 79.0 & 82.9 & 68.8 & 81.0 & 82.9 & 72.2 \\
        \midrule
        kNN & 82.5 & 83.9 & 75.3 & 84.4 & 85.4 & 78.1 \\
        AQE & 87.4 & 86.4 & 84.6 & 89.0 & 88.3 & 86.9 \\
        $\alpha$QE & 87.3 & 86.2 & 84.4 & 89.1 & 88.6 & 86.9 \\
        RDP & 88.9 & 88.1 & 86.7 & 91.3 & 90.3 & 89.7 \\
        SCA & 90.6 & 88.1 & 89.1 & 92.6 & 89.9 & 91.9 \\
        $k$-recip & 90.2 & 88.9 & 89.2 & 92.2 & 90.6 & 91.8 \\
        ECN & 90.3 & 88.9 & 89.5 & 92.5 & 90.9 & 92.1 \\
        \midrule
        Ours & \textbf{91.5} & \textbf{90.3} & \textbf{90.7} & \textbf{93.2} & \textbf{91.9} & \textbf{93.0} \\
        \bottomrule 
        \end{tabular}}
    \end{minipage}
    
\end{table}

\begin{figure}[t]
    \centering
    \subfigure[Ablation studies of $\kappa$.]{
    \begin{minipage}[b]{0.4\textwidth}
        \includegraphics[width=1\textwidth]{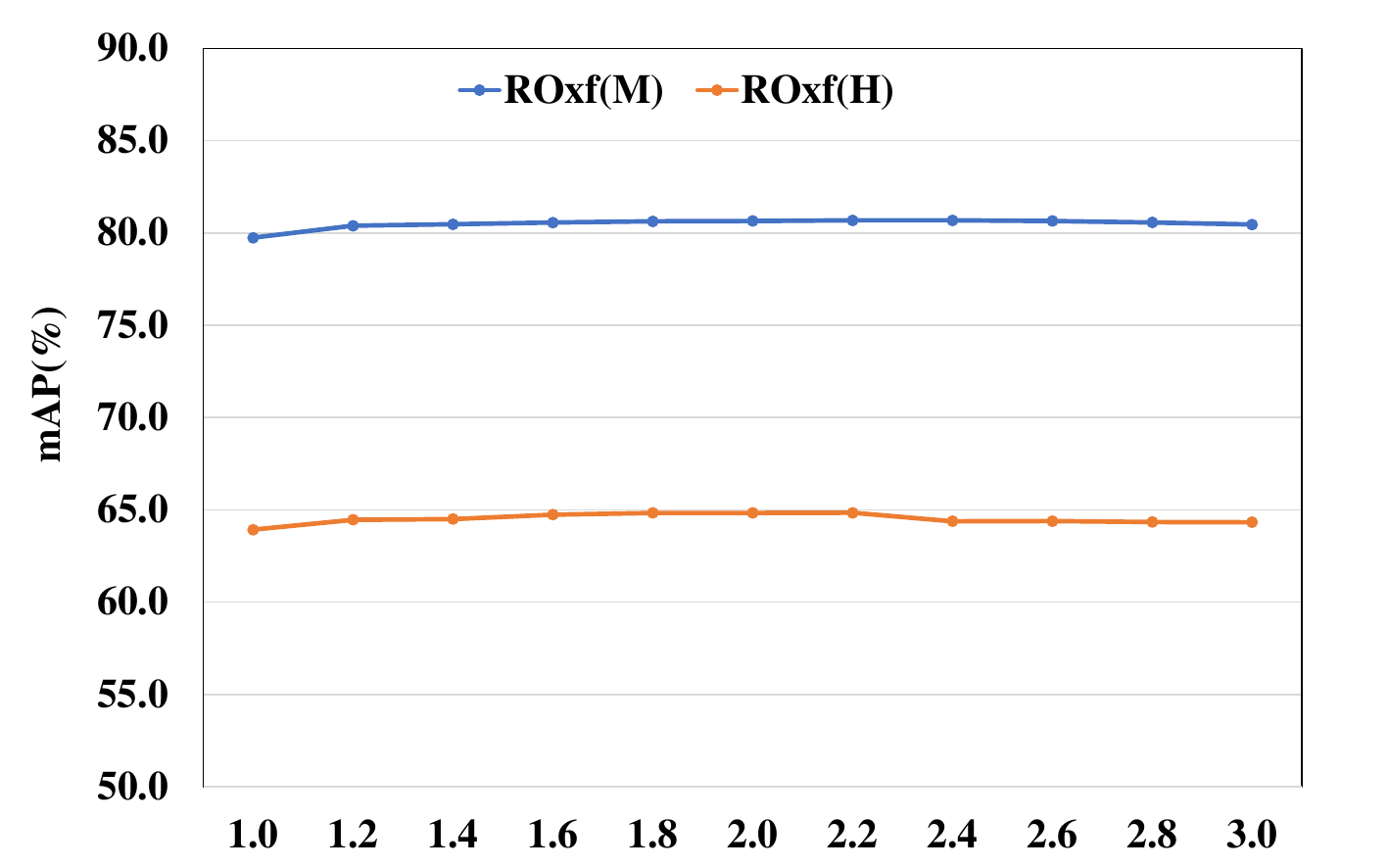}
        \label{fig: NSS ablation kappa}
    \end{minipage}}
    \centering
    \subfigure[Ablation studies of $\sigma$.]{
    \begin{minipage}[b]{0.4\textwidth}
        \includegraphics[width=1\textwidth]{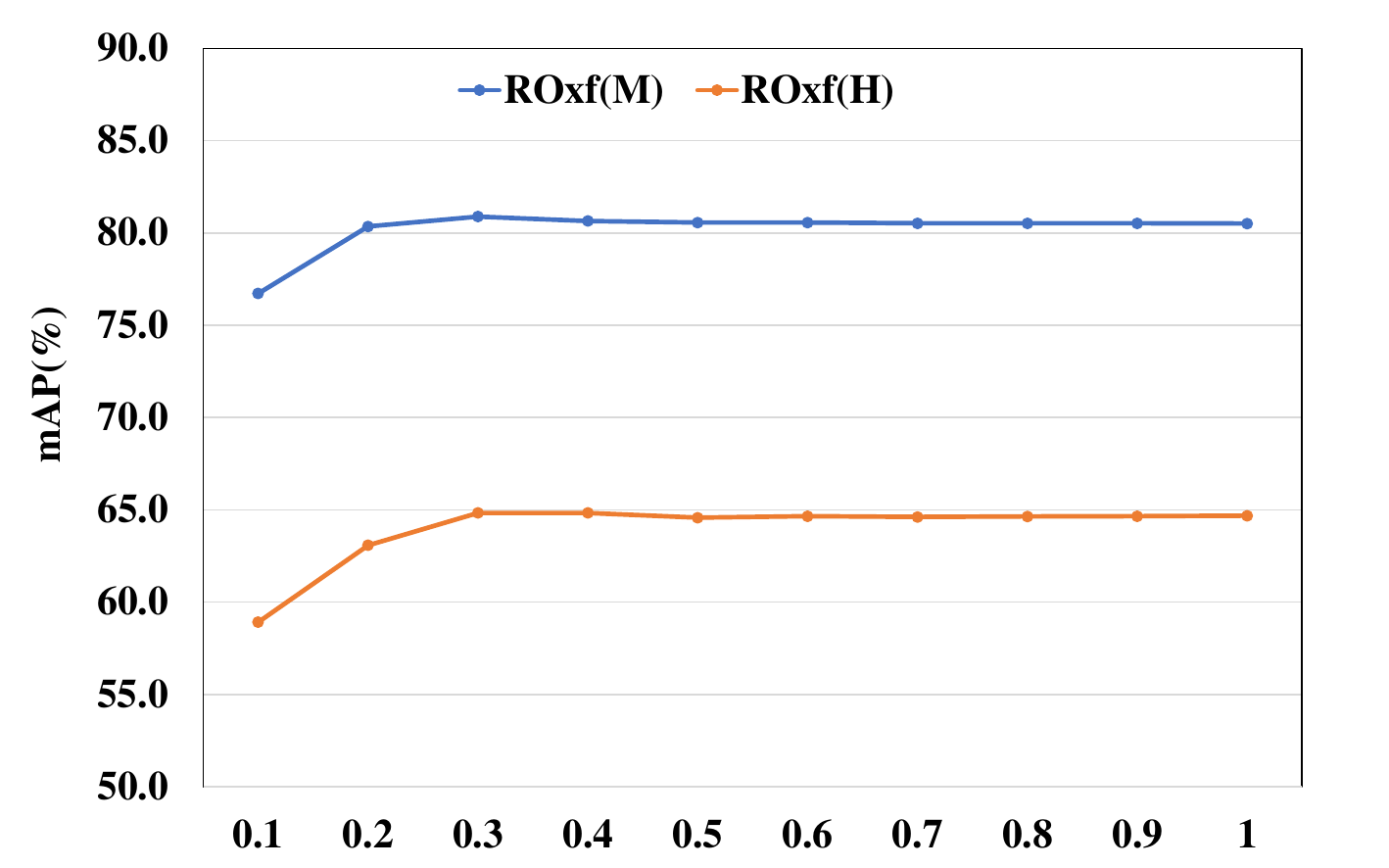}
        \label{fig: NSS ablation sigma}
    \end{minipage}}
    
    \subfigure[Ablation studies of $\mu$.]{
    \begin{minipage}[b]{0.4\textwidth}
        \includegraphics[width=1\textwidth]{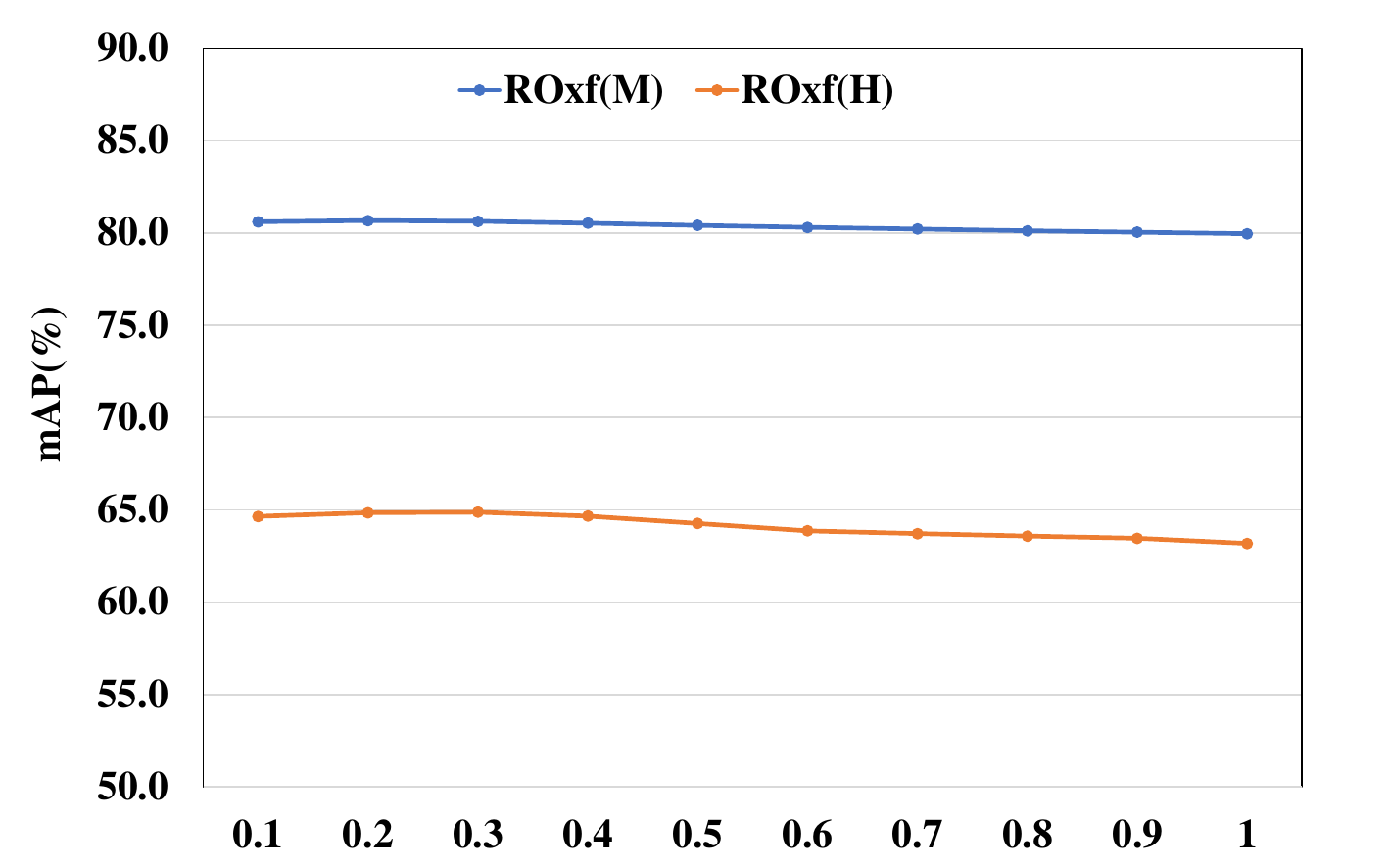}
        \label{fig: NSS ablation mu}
    \end{minipage}}
    \centering
    \subfigure[Ablation studies of $\beta$.]{
    \begin{minipage}[b]{0.4\textwidth}
        \includegraphics[width=1\textwidth]{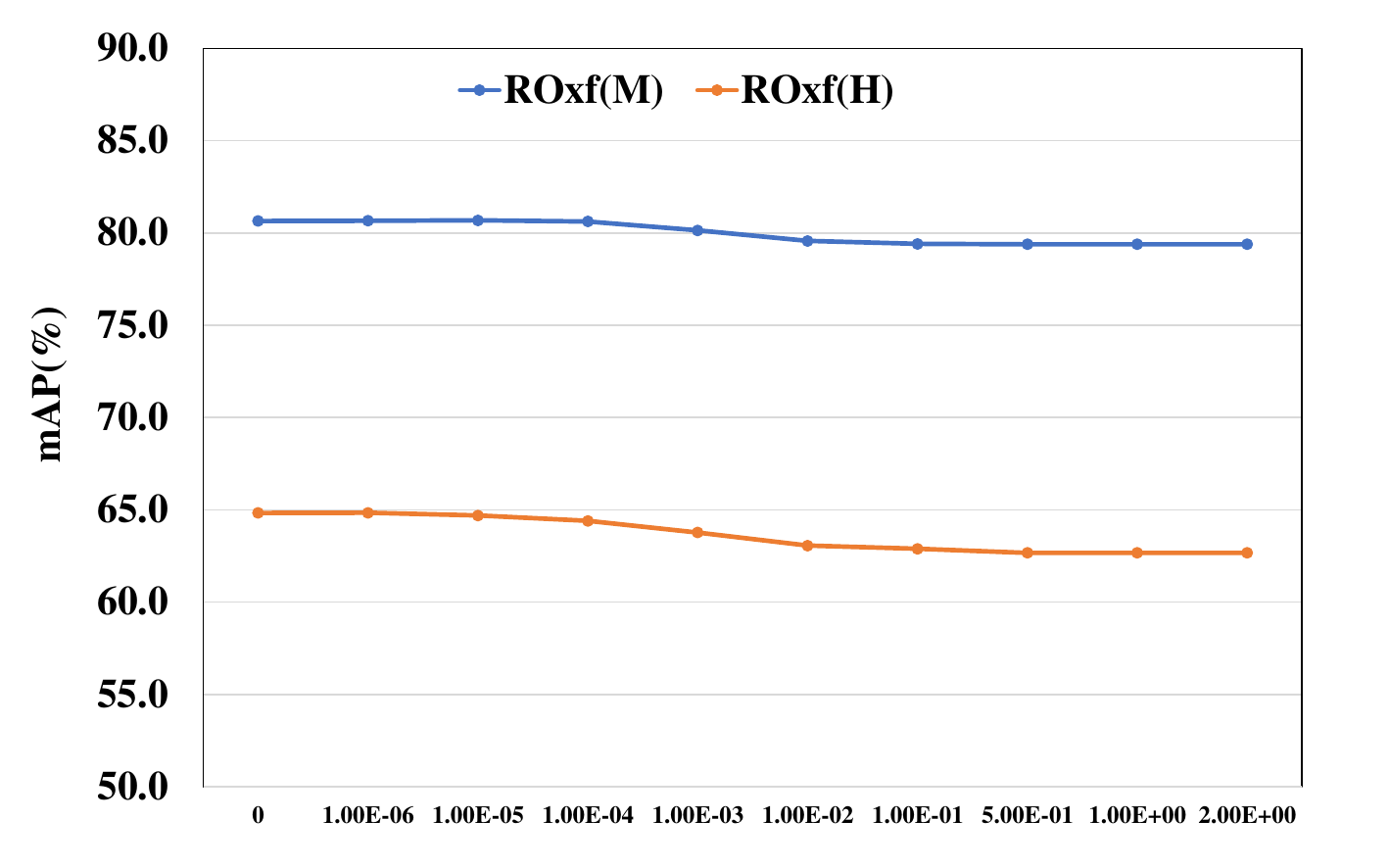}
        \label{fig: NSS ablation beta}
    \end{minipage}}  
    \caption{Extended ablation studies of other hyper-parameters. Test with R-GeM \cite{tpami2019_rtc} extracted descriptors.} 
    \label{fig: appendix ablation NSS}
\end{figure}


\end{document}